\documentclass[10pt,twocolumn,letterpaper]{IEEEtran}
\usepackage{times}
\usepackage{epsfig}
\usepackage{graphicx}
\usepackage{amsmath}
\usepackage{amssymb}
\usepackage{amsfonts}
\usepackage{amsthm}
\usepackage{units}

\begin{document}

\title{Partial Light Field Tomographic Reconstruction From a Fixed-Camera Focal Stack}

\author{Antoine Mousnier, Elif Vural, and
Christine Guillemot\\
INRIA, Campus Universitaire de Beaulieu, 35042 RENNES, FRANCE
\\
{\tt\small firstname.lastname@inria.fr}
}

\maketitle

\begin{abstract}
This paper describes a novel approach to partially reconstruct high-resolution 4D light fields from a stack of differently focused photographs taken with a fixed camera. First, a focus map is calculated from this stack using a simple approach combining gradient detection and region expansion with graph cut. Then, this focus map is converted into a depth map thanks to the calibration of the camera. We proceed after this with the tomographic reconstruction of the epipolar images by back-projecting the focused regions of the scene only. We call it masked back-projection. The angles of back-projection are calculated from the depth map. Thanks to the high angular resolution we achieve by suitably exploiting the image content captured over a large interval of focus distances, we are able to render puzzling perspective shifts although the original photographs were taken from a single fixed camera at a fixed position. 
\end{abstract}

\section{Introduction}
\label{sec:introduction}

A light field is a 4-dimensional function that measures the light along each ray reaching the camera sensors and not only the sum of the light rays striking each point in the image as in traditional cameras.  The recorded flow of rays allows the simulation of an image capture at different focal distances (digital refocusing) and different depths of field (digitally extended depth of field), the simulation of lenses with large aperture (synthetic aperture), and more generally, allows flexible image manipulation.
Several camera architectures have been proposed to capture the light field signal by inserting either an array of microlenses \cite{Ng} or a mask \cite{Veera} in front of the photosensor. 
Other solutions based on coded aperture \cite{Liang} or camera array \cite{Wilburn} designs have also been developed, but unlike the microlens array design which captures the light field in one shot, the latter solutions require the capture of multiple images with a bulky set-up to attain a sufficient angular resolution. In addition, heterodyne mask-based cameras, and external arrangements of lenses and prisms present limitations, such as loss of light and poor optical performance. Compared to camera arrays, plenoptic cameras avoid problems of synchronization and calibration. A review of light field acquisition devices can be found in \cite{Wetzstein}. Note that compressive sensing has also been recently considered as a framework for light fields camera design \cite{Babacan2012}.

Light field cameras using arrays of microlenses in front of the photosensor are becoming commercially available. The Raytrix company proposes cameras for industrial and scientific applications since 2010 and the Lytro company released the first light field camera for consumer application in early 2012. The array of microlenses placed in front of the photosensor is used to separate the light rays striking each microlens, and to focus them on different sensors according to their directions. This is, however, achieved by trading-off angular resolution (related to the number of sensor pixels corresponding to a microlens) and spatial resolution (related to the number of microlenses). This trade-off  leads to a significantly lower spatial resolution compared to traditional 2D cameras. For instance, the first-generation LYTRO camera uses a $11$-MP sensor and a $325 \times 400$ microlense grid, which means that, without super-resolution, the output image is $0.1$ MP. With super-resolution, the output image resolution can be raised to $1.2$ MP. The recently released Lytro Illum (April 2014) yields $2$-MP images with its $40$-MP CCD sensor. This is still far from the 8-MP resolution offered by recent smart phones.

Various approaches have been proposed to cope with the problem of poor resolution in  light field photography. The first approach is to directly super-resolve the light field, which means to enhance the spatial resolution as well as the angular resolution \cite{Bishop}, \cite{Wanner}, leading, however, to high volumes of data. The second approach is to super-resolve the rendered images using the available information in the light field. However, since a $11$-megaray camera provides an output image of only 0.1 MP, one would need to multiply the resolution by $80$ to attain a smartphone quality. High-resolution light fields can be captured using camera arrays, however, such a bulky set-up is not suitable for handheld capturing devices. For static scenes, a moving camera can be a practical alternative to camera arrays \cite{Gortler}. The focal sweep camera described in \cite{Zhou} combines focal sweep and continuous shooting in order to speed up the capture, but it mainly focuses on time and focus navigation effects called ``breathing pictures".

In this paper, we describe an alternative approach that consists in partially reconstructing the light field from a small set of images taken by a fixed camera at different focuses. Note that the problem of reconstructing a 4D light field from a focal stack has been already addressed in \cite{Levin2010}. However, the authors in \cite{Levin2010} consider a different approach which places the focal stack spectra (2D slices) at entries $L(\omega_x, \omega_y, s\omega_x, s\omega_y)$ in the 4D light field spectrum, the rest of the entries being set to zero. The slope $s$ of the slices depends on the focus distance of the image in the focal stack. The spectra of all shifted focal stack images are averaged and deconvolved using a slope-invariant kernel. The authors in \cite{Kubota2007} describe a method for reconstructing a dense light field from an array of multifocus images captured by several cameras.

In contrast, our approach considers a set-up with one unique camera at a fixed position which captures images at different focuses, forming a focus stack. The proposed method builds on the observation that the ordinary photographs of the focal stack are 2D projections of the 4D light field \cite{Ng}. Therefore, if one has sufficiently many projections of the light field, one can recover it by employing the reconstruction techniques used in tomography. Moreover, unlike the medical volumes in tomography, the 4D light field to be reconstructed has a particular structure. Exploiting this structure, we show that it is possible to achieve very satisfactory reconstructions of the light field for practical purposes (refocus, extended focus and perspective shifts), even with a small number of projections (focal stack images).

The proposed method first requires recovering the depth map of the scene by identifying the focal stack image where each pixel is in focus.  Once the focus measure is obtained, the focal stack image where each pixel is the sharpest is determined, which is then used to obtain a depth map of the scene. Note that, although our method requires the estimation of the depth map of the scene, the main problem treated in this paper is the reconstruction of the light field using tomographic backprojection, rather than the estimation of a depth map. 

Several previous works have indeed addressed the problem of computing depth from focus clues, which can be categorized in two main groups as depth from focus and depth from defocus methods. The problem of computing a depth map based on the sharpness of each pixel in a sequence of images captured at different focus settings is generally referred to as depth from focus. The method in \cite{Nayar} computes a dense depth map by proposing a sum-modified-Laplacian (SML) focus measure and extracting the depth information from the evolution of the focus measure between consequent images of a focal stack. In \cite{Asada}, a method is proposed to jointly detect edges and estimate depth information from a set of multi-focus images, while the depth is recovered only along the edges. An optical focus measure to obtain a depth map from a focal stack is proposed in \cite{MalikSC07}, where a band-pass filter to detect sharpness is designed based on bipolar incoherent image processing. The focus measure proposed in \cite{MalikSC11} is obtained by taking the SML focus measure \cite{Nayar} as an initial estimate and improving it by fitting a linear regression model. Some other focus measures  based on feature detection \cite{MendaparaMW09} and the complex wavelet transform \cite{MendaparaBW10} have also been proposed for depth estimation. Such focus measures rely in general on the presence of texture and their performance may be affected by the lack of texture. 

There is another category of methods known as depth from defocus which refers to the estimation of a depth map based on defocus clues. The depth from defocus method in \cite{Shaojie} computes a depth map by estimating the gradient magnitude ratio between the input image and a re-blurred version of it using a known Gaussian function, which is however designed for a single input image. Based on the observation that objects of interest are usually in-focus, the authors in \cite{Jiang} detect image salient regions by computing the degree of focusness, which measures the spread or scale of edges. The methods in \cite{Tao} and \cite{Kim} exploit both defocus and correspondence cues in order to build a depth map from light field data.

To be used in the proposed masked back-projection algorithm for tomographic reconstruction, the computed depth map should be dense and sufficiently reliable even in image regions with weak texture. The denseness of the depth map is of more critical importance in our problem than its precision, contrary to the main focus of some other works such as \cite{Nayar}.  For this reason, we propose a method that first searches for the sharpest regions in each photograph of the stack with a simple algorithm based on thresholded image gradients, which are assumed to give the in-focus zones. 
To get around the problem of lack of texture in some regions, we apply an additional stage of diffusing the focus map via graph cuts.
More precisely, we perform a region growing process with a graph cut algorithm to jointly estimate the complete focus map from all images in the stack, which permits the diffusion of the focus information in reliable (high-texture) regions to the regions where the sharpness estimate is less reliable. The focus map is then turned into a depth map by calibrating the camera. This procedure gives a depth map whose precision is limited to the number of focal stack images as it is based on assigning an image index to each pixel; however, it is dense due to the diffusion with graph cuts. Besides allowing the reconstruction of the light field, the estimated focus map can be directly used in obtaining an all-in-focus image (extended depth of field). We note that the problem of extended depth of field has also been addressed, e.g.~in \cite{Zhou} where the authors propose a focal sweep imaging system which captures a focal stack by physically sweeping the focal plane across the scene. Unlike the focal sweep camera of \cite{Zhou}, the imaging system proposed in \cite{CAVE} controls the depth of field by varying the detector position during the integration time of a single image.

Once we compute the focus and depth maps, we apply a masked back-projection algorithm for a tomographic reconstruction of the epipolar images, which are slices of the 4D light field. This masked back-projection algorithm is adapted to the particular structure of the epipolar images. The reconstruction of epipolar images enables advanced image manipulations such as perspective shifts, refocusing and the rendering of extended focus images. 

The proposed method for light field reconstruction has several advantages over the previous light field capturing solutions. First, it makes it possible to capture the light field with any conventional 2D camera, hence, disposing of the need for specially devised cameras. Next, as the camera is held at a fixed position, it does not require any additional procedures such as camera pose estimation. Moreover, unlike the previous light field capturing mechanisms, the proposed light field reconstruction method does not impose any limitations on the image resolution, simply because the resolution of the rendered images are determined by the resolution of the 2D camera used in capturing the focal stack.
Compared to the method in \cite{Levin2010}, which also aims to reconstruct light fields from focal stacks, the proposed method is experimentally observed to be more robust to variations in image capturing conditions, such as the number of images in the focal stack and the depth ranges of the scene and of the focal stack.

The rest of the paper is organized as follows. In Section \ref{sec:background}, we give an overview of light field photography. In Section \ref{sec:algorithm}, we present our method for tomography-based 
 light field reconstruction. In Section \ref{sec:experiments}, we demonstrate the application of our algorithm in depth map estimation, extended focus and perspective shift problems. Finally, we conclude in Section \ref{sec:conclusion}.

\section{Light Fields: A Review}
\label{sec:background}

The plenoptic function is a 7D function $L(x,y,z,\theta, \phi, t, \lambda)$ describing the radiance (intensity) of the light rays emitted by an object (or a scene) and received by an observer at a point $(x,y,z)$ in space, along a direction of gaze $(\theta, \phi)$, at a time instant $t$ and wavelength $\lambda$ \cite{Adelson}. From the plenoptic function, one can reconstruct every possible view at every time instant, from every position, and at every wavelength. 
If the time parameter (the scene) is assumed fixed and the wavelength parameter is conventionally sampled and reconstructed at three positions (red, blue and green), we are left with a 5D function. The light ray in the 3D space is in this case represented by the coordinates $(x,y,z)$ of the 3D point the light ray passes through and by its direction $(\theta, \phi)$.

\subsection{Two-plane parameterization}
However, the radiance along a ray emitted from a convex object is  constant if the light ray does not encounter an occluding object. The dimension of the plenoptic function can therefore be decreased by one, resulting in a 4D function called light field \cite{Levoy} or Lumigraph \cite{Gortler}. 
The light field can hence be parameterized with two bounded planes \cite{Levoy}. Every captured ray can be described by its intersection with the main lens (i.e., the first plane, parameterized by (${u,v}$)) and the sensors (i.e., the second plane, parameterized by (${x,y}$)). The image formation process from the light field, taking into account light ray propagation, lens refraction, occlusion, is modelled in \cite{Liang2011}.
The light field representation also helps understanding various computational camera designs \cite{Zhou2011}.

In conventional cameras, each pixel is the integration of all the light rays that hit the corresponding sensor, yielding an integration over $(u,v)$. On the contrary, plenoptic cameras replace the sensors by a grid of microlenses and add sensors behind every microlens. Therefore, rays coming from different directions and falling onto the same microlens hit different sensors, which makes it possible to retrieve the part of the main lens hit by a given light ray. Every ray can be described by its intersection with the main lens and the grid of microlenses \cite{Ng}.

\begin{figure}[t]
\centering
\resizebox*{7cm}{!}{\includegraphics{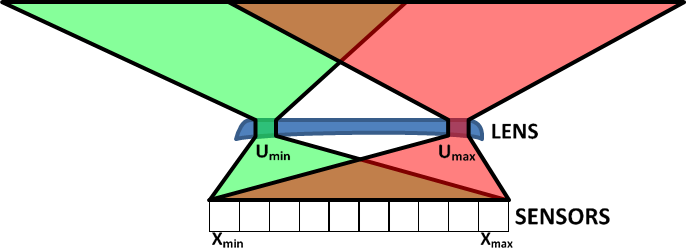}}
\caption{Sub-aperture selected by $u$ changes the angle of view.}
\label{apertures}
\vspace{-10pt}
\end{figure}

The recorded 4D light fields can be seen as capturing an array of viewpoints (sub-aperture images) of the imaged scene which are spread over the extent of the lens aperture.  As changing the sub-aperture means changing the point of view, objects are not projected on the same sensors, as schematized in  Fig.\ref{apertures}. These sub-aperture images give information about the parallax and depth of the imaged scene.

\begin{figure}[t]
\centering
\resizebox*{4cm}{!}{\includegraphics{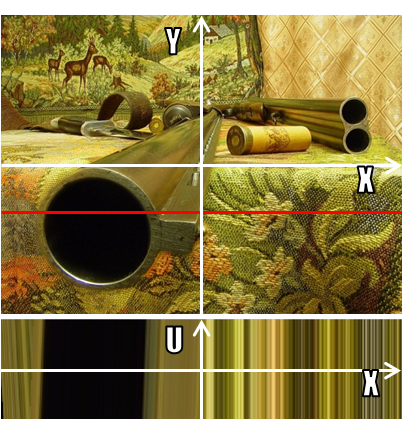}}
\caption{Observed line (in red in top image) and corresponding $(x,u)$ epipolar image (bottom image). Top image from \cite{Makaruk}.}
\label{EPI}
\vspace{-10pt}
\end{figure}

\begin{figure}[t]
\centering
\resizebox*{6cm}{!}{\includegraphics{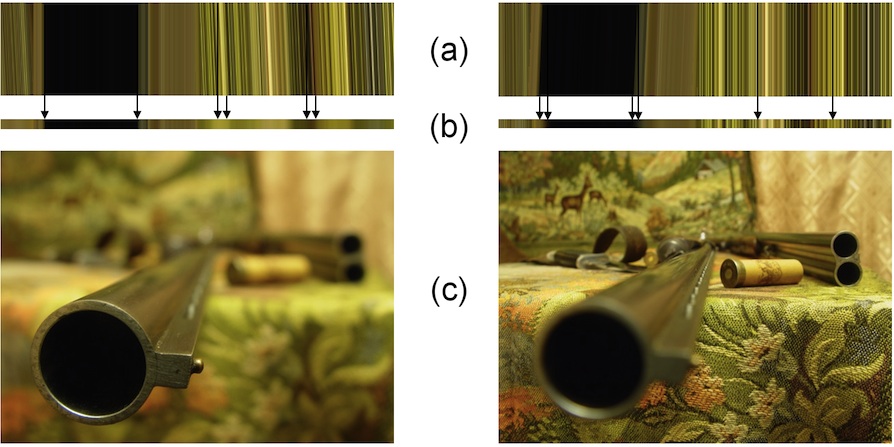}}
\caption{The focusing process. Both epipolar images in (a) correspond to the red line shown in Fig.~\ref{EPI}, but the left image is obtained by focusing on the closest part of the rifle whereas the right image is obtained by focusing on the tablecloth in the right lower part of the scene. The lines corresponding to focused zones are vertical, whereas the out-of-focus zones lines are inclined. The vertical integration of the epipolar images yields the lines in (b). The integration of vertical lines results in sharp edges, whereas the integration of inclined lines results in blurred edges. If we apply the same process to the whole image, we obtain (c).}
\label{integ}
\vspace{-10pt}
\end{figure}

\subsection{Epipolar images}

Epipolar planes are the two-dimensional $(x,u)$, $(x,v)$, $(y,u)$, and $(y,v)$ slices of the 4D light field. In the following, we only consider the $(x,u)$ slices. Fig.~\ref{EPI} shows a photograph extracted from the light field as well as one epipolar image, which is a bounded subset of an epipolar plane captured by a plenoptic camera. The epipolar image represents an $(x,u)$-slice of the light field and gives an observation of the light field at the constant $y$-value corresponding to the red line in the photograph. Each vertical line on the epipolar image represents the light field observed at varying sub-apertures $(u)$ of the main lens at a constant pixel location $x$. One can observe linear patterns of varying slope in the epipolar image. Each one of these lines corresponds to a certain scene point and is formed by the observation of this scene point with a continuous variation of the view angle. The slope of a line is then determined by the depth of the scene point it corresponds to. Points belonging to the focal plane of the optical system trace vertical lines, and the further a point is from the focal plane, the steeper is the slope of the line it traces in the epipolar image.

The capture of a photograph is the vertical integration of the epipolar images. Focusing on a point is equivalent to rotating the light field such that the line that corresponds to this point in the epipolar image becomes vertical. Consequently, the vertical integration (the capture) converts vertical lines into in-focus pixels, whereas non-vertical lines yield blurred pixels. Fig.~\ref{integ} illustrates this.

Digital refocusing can be seen as post-capture integration of light rays. Instead of integrating the epipolar image at the time of capture, a plenoptic camera records it in order to be able to integrate it afterwards, when the user selects the part of the scene to focus on. Refocusing on a zone is integrating the recorded light field in the direction given by the lines of the epipolar images in this zone.

From an algorithmic point of view though, it is more interesting to evaluate the integral in the Fourier domain than in the spatial domain, since integrating in the spatial domain is equivalent to taking a slice in the Fourier domain. Due to the Fourier slice theorem, this integral can in fact be evaluated by computing the 4D Fourier transform of the light field, taking the 2D slice passing through the origin perpendicularly to the given direction, and performing the inverse Fourier transform \cite{Ng}. As the distance between the desired focal depth and the reference focal distance used in the original image capture increases, the slope of the slice also increases.

\subsection{Limitations of plenoptic cameras}

\begin{figure}[t]
\centering
\resizebox*{6cm}{!}{\includegraphics{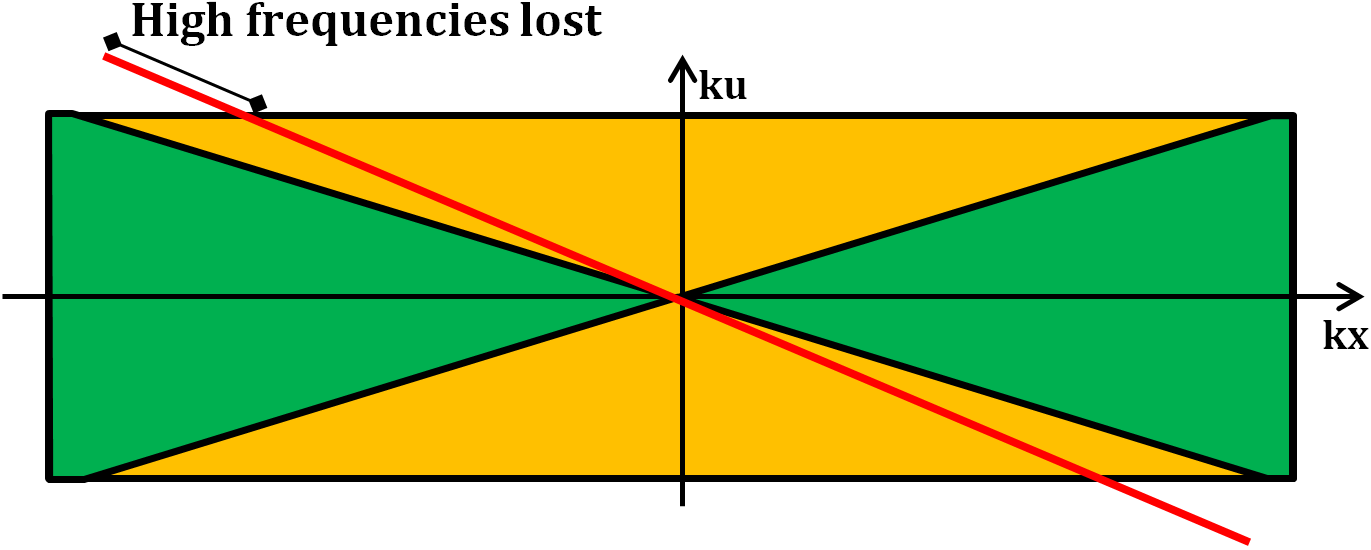}}
\caption{Loss of high frequencies when refocusing too close. Green regions correspond to the range of slices where exact refocusing is possible, whereas orange regions correspond to the range of slices where refocusing involves a loss of high frequencies.}
\label{LFlimit}
\vspace{-10pt}
\end{figure}

Plenoptic cameras with a grid of microlenses such as Lytro or Raytrix have limitations in terms of achievable resolution.  Refocused photographs have a smaller resolution than the originally focused image, and the resolution further decreases when refocusing too close or too far. This is illustrated in Fig.~\ref{LFlimit}. The first step of digital refocusing is the extraction of a 2D slice of the 4D space that passes through the origin. If the angle made by the 2D slice is too large, its portion corresponding to high spatial frequencies is limited (see Fig. \ref{LFlimit}). 

The resolution of the originally focused photograph depends only on the spatial resolution of the light field. Meanwhile, the resolution of a refocused photograph depends on the spatial resolution of the light field, its angular resolution, and the difference between the original focal distance and the new focal distance. From Fig.~\ref{LFlimit}, the following formula is obtained using basic geometric relations
\begin{equation}
    N_{Samples}=
          \left \{
   \begin{array}{l r}
	X_{res} \sqrt{1+(\frac{1-\alpha}{\alpha})^2}  & \text{ if  } \alpha\geq\frac{X_{res}}	{X_{res}+\Theta_{res}} \\
	\Theta_{res} \sqrt{1+(\frac{\alpha}{1-\alpha})^2} &  \text{ if  }  \alpha\leq\frac{X_{res}}{X_{res}+\Theta_{res}} \\
   \end{array}
   \right .
\end{equation}
where $N_{samples}$ is the spatial resolution of the refocused photograph, $X_{res}$ is the spatial resolution of the plenoptic camera, $\Theta_{res}$ is its angular resolution, $\alpha$ is the refocusing parameter \cite{Ng}. Thus, in order to increase the resolution of all refocused photographs by a factor of $N$, the number of sensors should be multiplied by $N^2$, as $X_{res}$ and $\theta_{res}$ both have to be multiplied by $N$.

\section{Light field reconstruction algorithm}
\label{sec:algorithm}

Since a photograph is the 2D projection of the 4D light field along the direction given by its focused depth \cite{Ng}, one can reconstruct the epipolar images from a focal stack. In a setting where a refocused image is constructed line by line, the first step is to reconstruct the 2D epipolar image corresponding to each horizontal line in the image plane. The problem of reconstructing a 2D epipolar image  is similar to the tomographic reconstruction in medical imaging, where a 2D image is reconstructed from its 1D projections. 

However, there are two major differences between classical tomographic reconstruction and our approach. First, the angles of projection are known in the case of medical tomography, since these angles can be measured directly on the X-ray emitting device. On the contrary, the angles of projection are not immediately known when reconstructing an epipolar image from a focal stack and thus need to be estimated. Second, while the reconstruction of the 2D complex and irregular textures of organs and tissues with the back-projection of 1D projections is a challenging problem, the back-projection technique provides a very convenient tool for the purpose of reconstructing 2D epipolar images, as these have a rather regular structure consisting of linear patterns of varying slope. The main problem is then to prevent lines from overlapping.

Let $LF(x,y,u,v)$ represent the 4D light field. The subset of the 4D light field that we aim to recover consists of the set {$LF(x,y,u,0)$}, which is sufficient to achieve image manipulations. In order to reconstruct this set with the back-projection technique, the angle of back-projection should be known for every pixel of the image. As discussed in Section \ref{ssec:epi_im_recon}, these angles can be derived from the depth map of the scene. We thus begin with estimating the depth map of the scene by first obtaining a focus map, and then transforming it into a depth map in meters thanks to the prior calibration of the camera (see Appendix).

\subsection{Focus and depth map estimation}
\label{ssec:DMestim}

\begin{figure}[t]
\centering
\resizebox*{6cm}{!}{\includegraphics{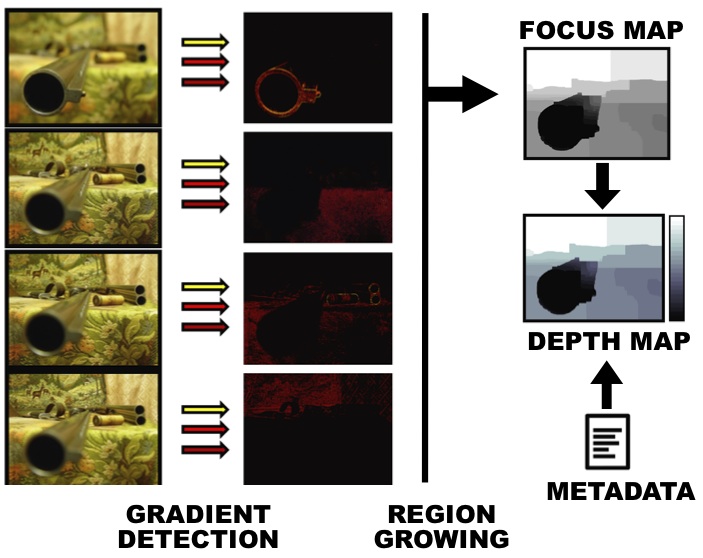}}
\caption{We first detect focused zones in each image with gradient thresholding (bright areas correspond to strong gradients). Next, we obtain a focus map by region growing, which gives the index of the image where each pixel is in-focus. The focus map is then converted into a depth map by using the camera metadata.}
\label{process}
\vspace{-10pt}
\end{figure}

We propose a two-step algorithm for focus map estimation. We first detect the strong gradients in each image of the focal stack and then use a graph cut algorithm to expand the zones where strong gradients have been detected. The process is described in Fig.~\ref{process}.

When the depth of field is shallow, focused zones are sharper than out-of-focus zones. One can thus assume that image intensity gradients are stronger in the focused regions of an image. Given a focal stack $\{ I_k\}_{k=1}^K$ of $K$ images, for each image $I_k$ in the focal stack, we build a layer $F_k$ that gives the ``sharpness scores'' of the pixels in $I_k$
\begin{equation}
F_k(x,y) = \sum_{m=1}^M \alpha_m u( \| \nabla I_k(x,y) \| - m \delta_k )
\end{equation}
where $u(\cdot)$ is the unit step function, $\nabla I_k (x,y) $ is the image intensity gradient at $I_k(x,y)$, and $\delta_k$ is the gradient threshold used for $I_k$ (selected automatically by a standard edge detection algorithm). The sharpness score $F_k(x,y)$ is thus obtained by comparing the image gradient value with a sequence of increasing thresholds $m \delta_k$, weighted with coefficients $\alpha_m$.

We then use the graph cut algorithm in \cite{Boykov} to diffuse the indices of the images where the pixels have the strongest gradients. Denoting the pixels as $X=(x,y)$, we compute a focus map $l$ such that $l_X:=l(x,y)$ gives the index of the focal stack image where the pixel $X$ is in focus.  The focus map $l$ is computed by minimizing an objective function $E(l)$ of the following form 
\begin{equation}
\label{eq:costfun_graphcut}
E(l)= \sum_{X} D(X, l_X) + \sum_{X \sim X'} S(l_X, l_{X'}).
\end{equation}
Here $D(X, l_X)$ is a data term that represents the cost of assigning the label $l_X$ to the pixel $X$. The data cost $D$ is determined with respect to the sharpness scores of the pixels as
\begin{equation}
D(X,l_X) = \big(\sum_{m=1}^M \alpha_m \big) - F_{l_X} (x,y).
\end{equation}
Hence, the data cost $D$ takes nonnegative values that decrease with the sharpness scores of the pixels.

The second term in (\ref{eq:costfun_graphcut}) controls the smoothness of the variation of labels among neighboring pixels $X$ and $X'$ (denoted as $X \sim X'$). The summation runs over all pairs of neighboring pixels, and the smoothness term $S(l_X, l_{X'})$ is the cost of assigning the labels $l_X$ and $l_{X'}$ to $X$ and $X'$. We set the smoothness cost as 
\begin{equation}
    S(l_X, l_{X'})= \left \{
   \begin{array}{l r}
   0 \quad \text{ if } \, \, l_X = l_{X'}\\
   0.5 + \frac{\log(|l_X - l_{X'} |)}{\log K }  \quad \text{ if } \, \, l_X \neq l_{X'} \\
   \end{array}
   \right .
\end{equation}
where $K$ is the number of focal stack images. The smoothness cost increases with the label difference among neighboring pixels only at a logarithmic rate, in order to avoid the over-penalization of label differences along directions of depth discontinuity. The cost of the maximal label transition between adjacent pixels is thus bounded by three times that of a minimal one, by setting the smoothness term to take values between $0.5$ and $1.5$ when $l_X \neq l_{X'}$.


Minimizing the cost function $E(l)$, we thus obtain a focus map, which makes it possible to compute an all-in-focus image (see Section \ref{sec:experiments}). Using the camera calibration parameters and the image capture metadata, the focus map can be easily converted to a depth map (see Appendix).

\begin{figure}[t]
\centering
\resizebox*{5cm}{!}{\includegraphics{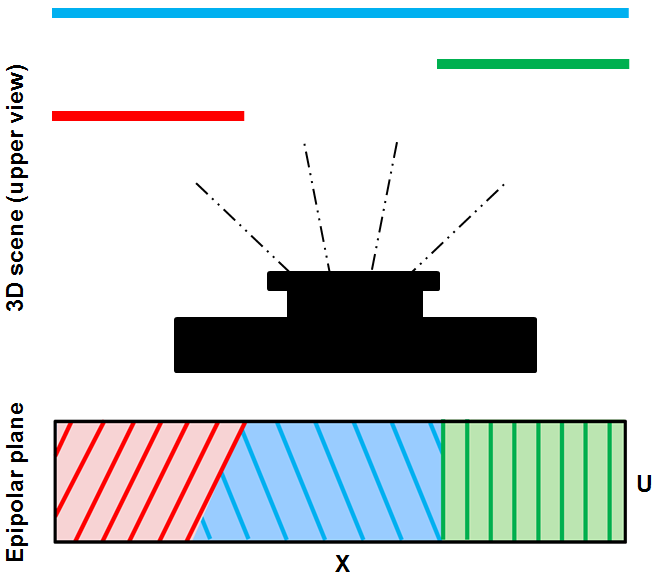}}
\caption{Simple scene and corresponding epipolar image. The green plane is the reference plane.}
\label{epipolar}
\vspace{-10pt}
\end{figure}

\subsection{Epipolar image reconstruction}
\label{ssec:epi_im_recon}

Based on the observation that a focal stack can be viewed as a collection of projections of the 4D light field along different directions, we propose to retrieve the epipolar images from their projections with an algorithm that we name masked back-projection. Consider the problem of retrieving an epipolar image for the simple scene illustrated in Fig.~\ref{epipolar} with only three different depth levels. Let the green plane be chosen to lie at the reference focal distance, so that the points on this plane correspond to vertical lines in the epipolar image.

\begin{figure}[t]
\centering
\resizebox*{5cm}{!}{\includegraphics{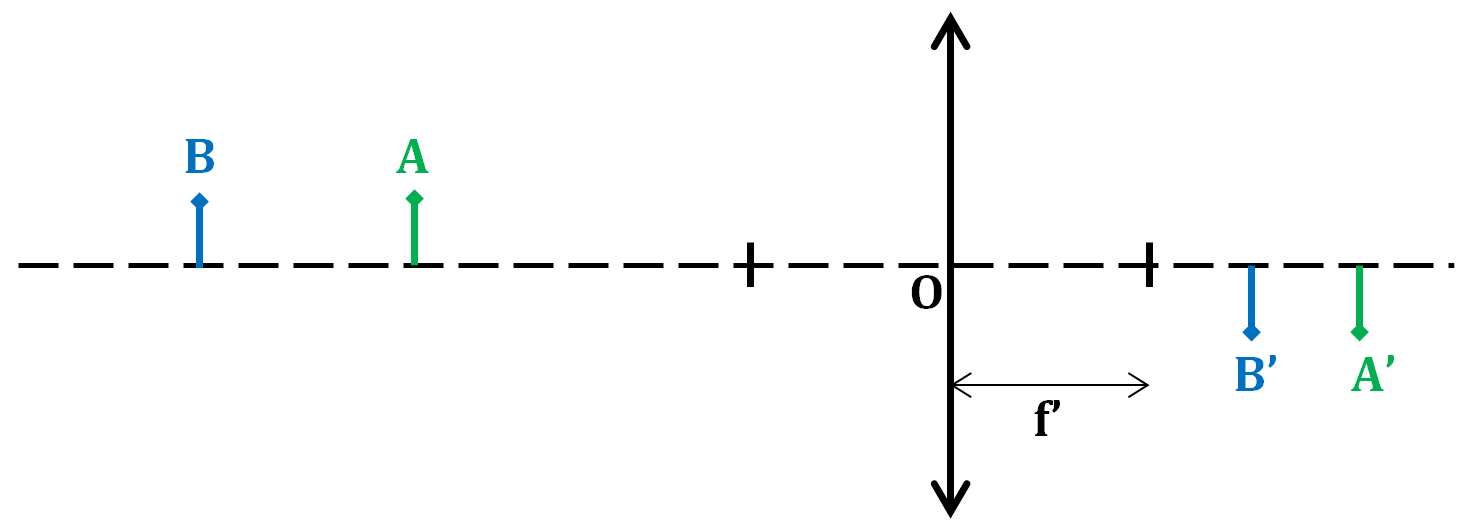}}
\caption{Optical diagram. $A'$ and $B'$ are respectively the images of $A$ and $B$ formed by the lens of center $O$ and focal length $f'$.}
\label{opticalDiagram}
\vspace{-10pt}
\end{figure}

\subsubsection{Angle of back-projection}

The back-projection angle at a pixel, i.e., the slope of its corresponding line in the epipolar image, can be estimated from the depth value of the pixel. Parameterizing the slope of a line with a vector $(x,u)$, the projection angle $\theta$ is given by
$
\tan{\theta} = u/x.
$
Fig.~\ref{opticalDiagram} is an optical diagram representing two points $A$ and $B$ (lying respectively on the green reference plane and on the blue plane) and their images $A'$ and $B'$. The positions of these images are the positions of the sensors when the camera is focused respectively on these two planes. The following equation from \cite{Ng} provides a relation between the components of the slope vector and the distances between the lens and the sensor positions
$
       u = (1/ \alpha -1 )x 
$
where $\alpha = ( \overline{OB'}) / (\overline{OA'})$.
Here, $\overline{OA'}$ and $\overline{OB'}$ respectively represent the distances between the lens centre $O$ and the sensor positions $A'$ and $B'$. These distances are positive, whereas $\overline{OA}$ and $\overline{OB}$ are negative.

The lens-object distance (object depth) and the lens-sensor distance are related through the thin lens equations as 
$
(\overline{OA'})^{-1} - (\overline{OA})^{-1} = (f')^{-1}
$
and
$
(\overline{OB'})^{-1} - (\overline{OB})^{-1} = (f')^{-1}.
$
From these equations, we can express the angle of projection $\theta$ as a function of the distance to the focused object $D = \overline{OB}$, the distance to the reference object $D_{ref} = \overline{OA}$, and the focal length of the lens $f'$.
\begin{equation}
\label{eq:proj}
\theta = \arctan{\frac    {\nicefrac{1}{D_{ref}}-\nicefrac{1}{D}}   {\nicefrac{1}{f'}-\nicefrac{1}{D_{ref}}}   }
\end{equation}

\subsubsection{Masked back-projection}

The conventional back-projection method is based on the Radon transform. 
The projections of the image along a set of different directions give its Radon transform. By back-projecting a filtered version of each projection along the corresponding angle and summing up the back-projections, one can retrieve the original image.


In order to achieve an accurate reconstruction with the traditional back-projection algorithm, a high number of projections is needed, whereas the number of projections is limited by the number of images in the focal stack in our problem. Nevertheless, the epipolar image to be reconstructed has a special structure, which can be approximated as a set of overlapping lines with different slopes. We thus propose a modified version of the back-projection algorithm adapted for our problem, which exploits this prior information on the epipolar image.

We back-project the pixels in each image with the corresponding projection angle estimated from the depth map, while we follow some rules for the registration of the back-projected pixels. 
We first back-project entirely the background. Then, we back-project only the in-focus parts of the projections, from the second farthest to the foreground, in the order of decreasing depth. An important difference between our back-projection procedure and the traditional one is that we overwrite the reconstructed epipolar image throughout the back-projections, while the classical method sums them up. The linear patterns in the epipolar image correspond to points on objects in a real 3D scene; therefore, processing the objects in the order of decreasing depth and overwriting the previous pixels throughout the back-projection is coherent with the physical fact that distant objects are occluded by the closer ones.


In the back-projection, we use masks to prevent the out-of-focus parts of the images from being back-projected, except for the background, for which we back-project every pixel. We thus determine the in-focus regions in each image in the focal stack, and build 3D blocks by back-projecting the pixels in these regions with the back-projection angle determined from the depths of the regions. The subset  $LF(x,y,u,0)$ of the light field is thus reconstructed progressively, by following the procedure described above. The subset $LF(x,y,0,v)$ can be reconstructed in a similar way, based on the $(y,v)$-epipolar images.

\section{Experimental Results}
\label{sec:experiments}

In our experiments we have used three types of data sets: 
(i) two data sets captured with a Nikon D5200 camera with an objective Nikon DX AF-S Nikkor 18:105 mm 1:3.5-5.6, (ii) the data set in \cite{Makaruk}, which contains 24 focal stack images, and (iii) a data set captured with a Lytro camera which is available for download at the following link:  http://www.irisa.fr/temics/demos/lightField/index.html. As the Nikon camera does not produce the ``distance to subject" metadata, we have used the DigicamControl software and calibrated the camera in order to determine the focal distance corresponding to each image in the focal stack, which is explained in the Appendix. In the following, we present results on the estimation of the depth map from a focal stack and demonstrate the usage of the reconstructed light field in extended focus and perspective shift applications. More results with also refocusing applications, and in particular the videos of the perspective shifts functionality can be accessed at the project page: http://www.irisa.fr/temics/demos/lightField/LightFieldPartialReconstruction.html

\subsection{Depth map estimation}

\begin{figure*}[t]
\centering\resizebox*{3cm}{!}{\includegraphics{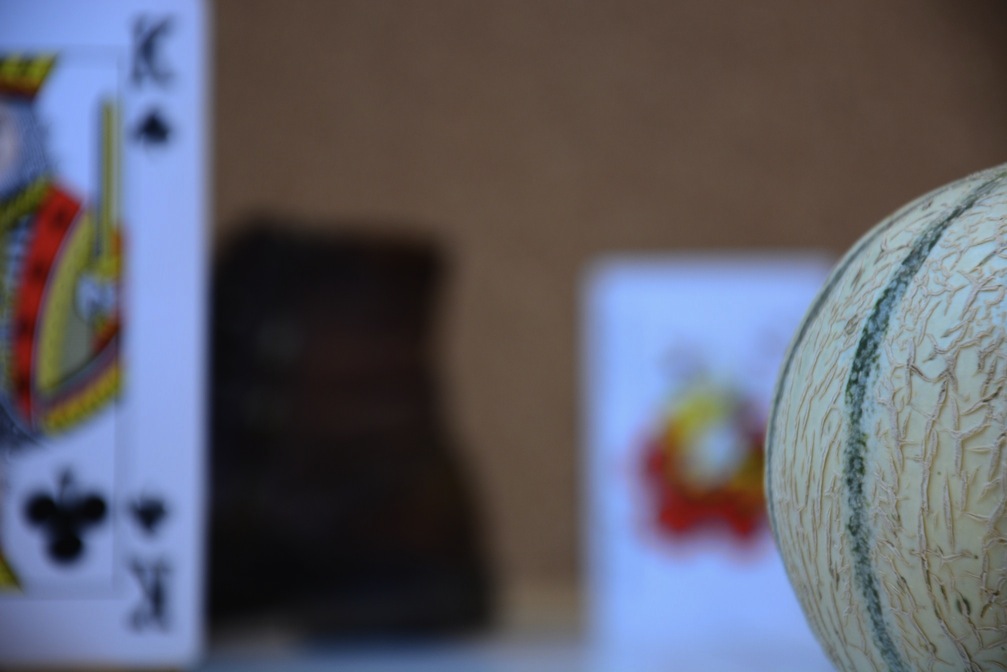}}
\centering\resizebox*{3cm}{!}{\includegraphics{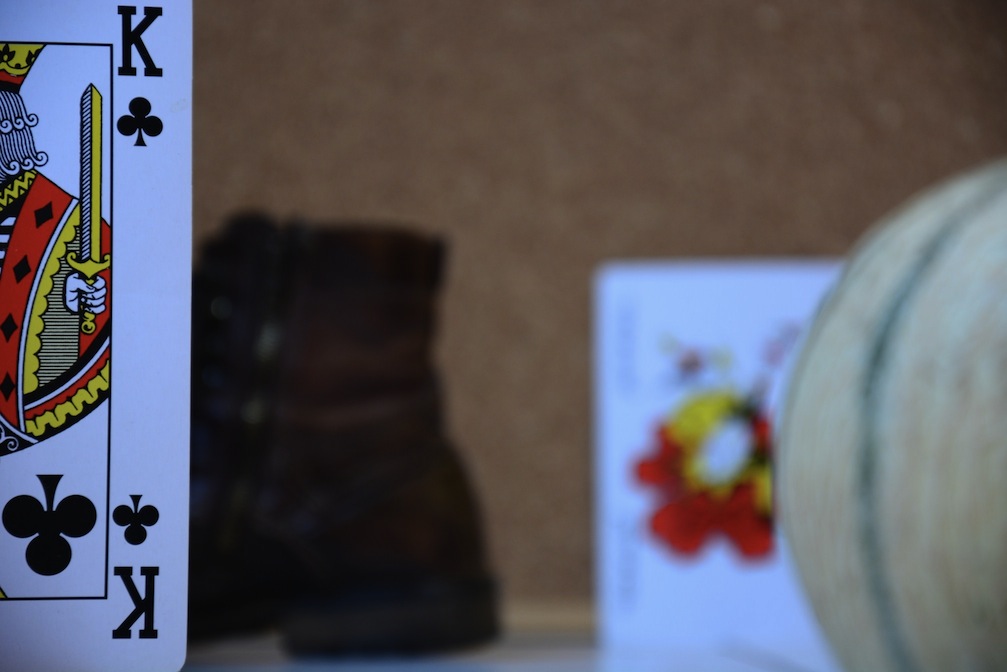}}
\centering\resizebox*{3cm}{!}{\includegraphics{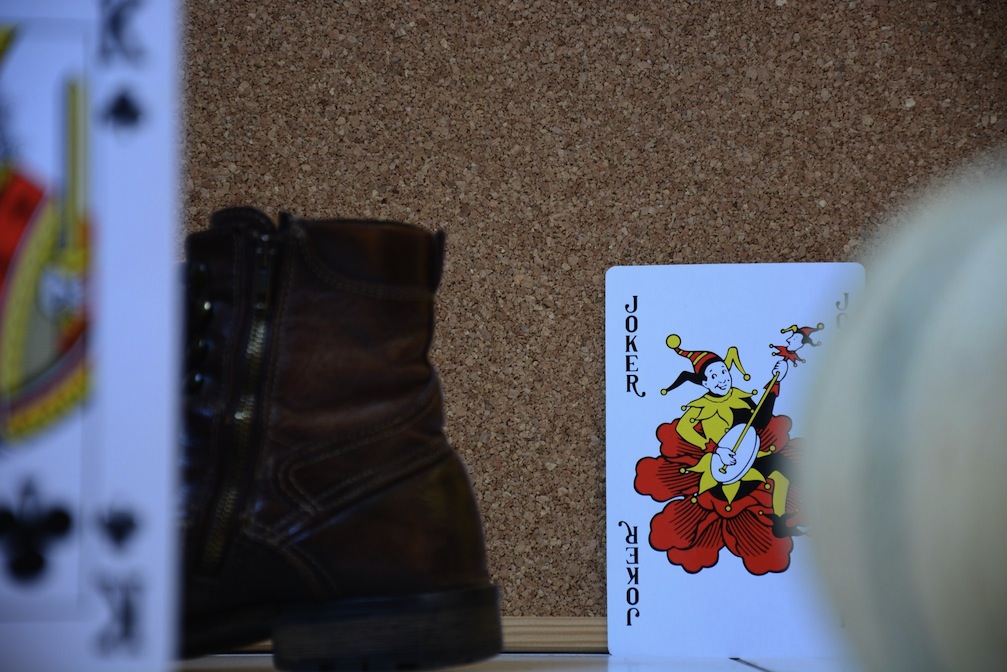}}
\centering\resizebox*{3.2cm}{!}{\includegraphics{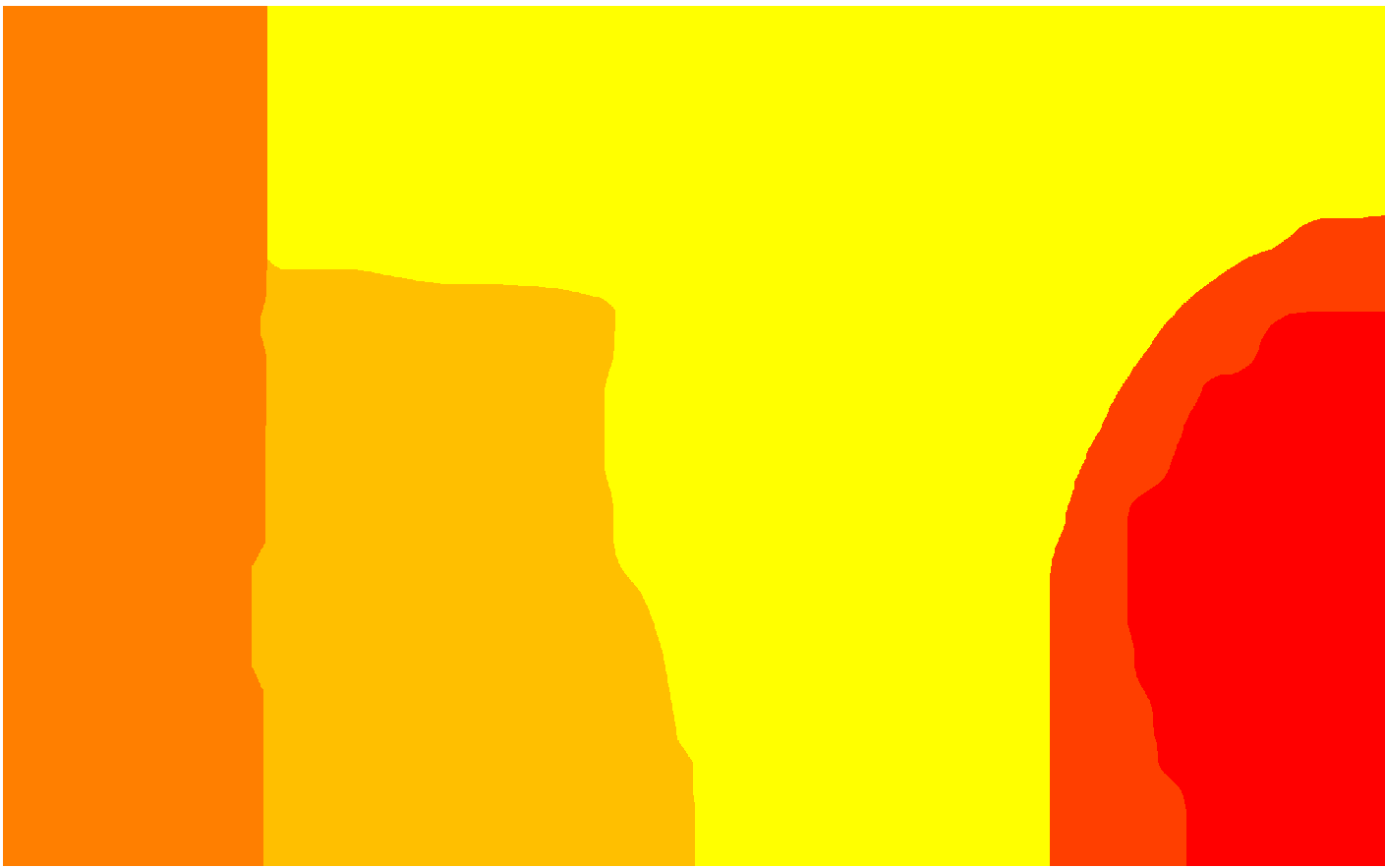}}
\centering\resizebox*{3cm}{!}{\includegraphics{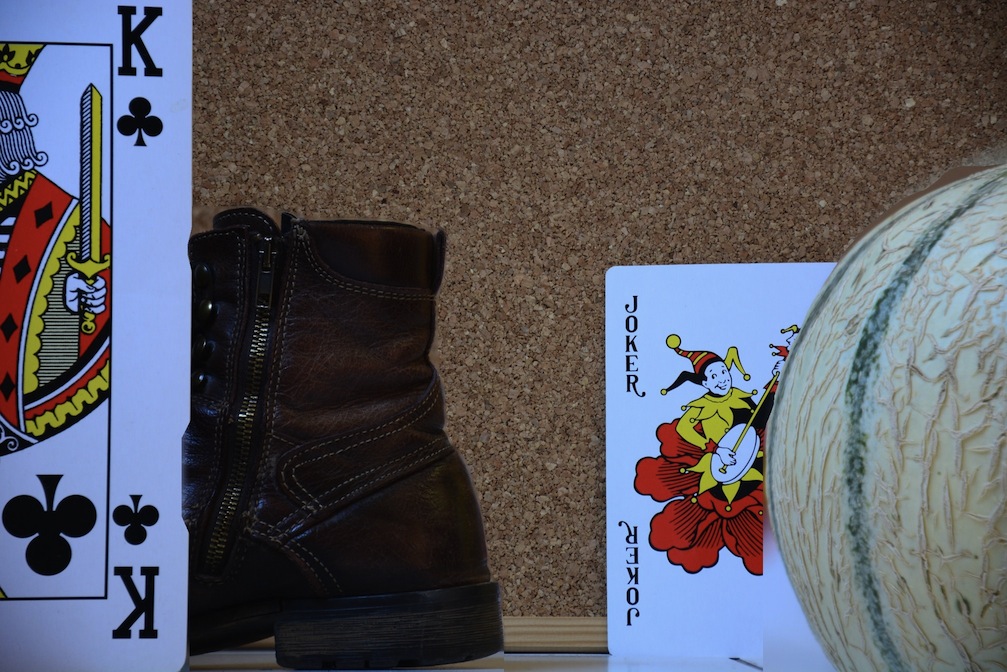}}\\
\vspace{0.1cm}
\centering\resizebox*{3cm}{!}{\includegraphics{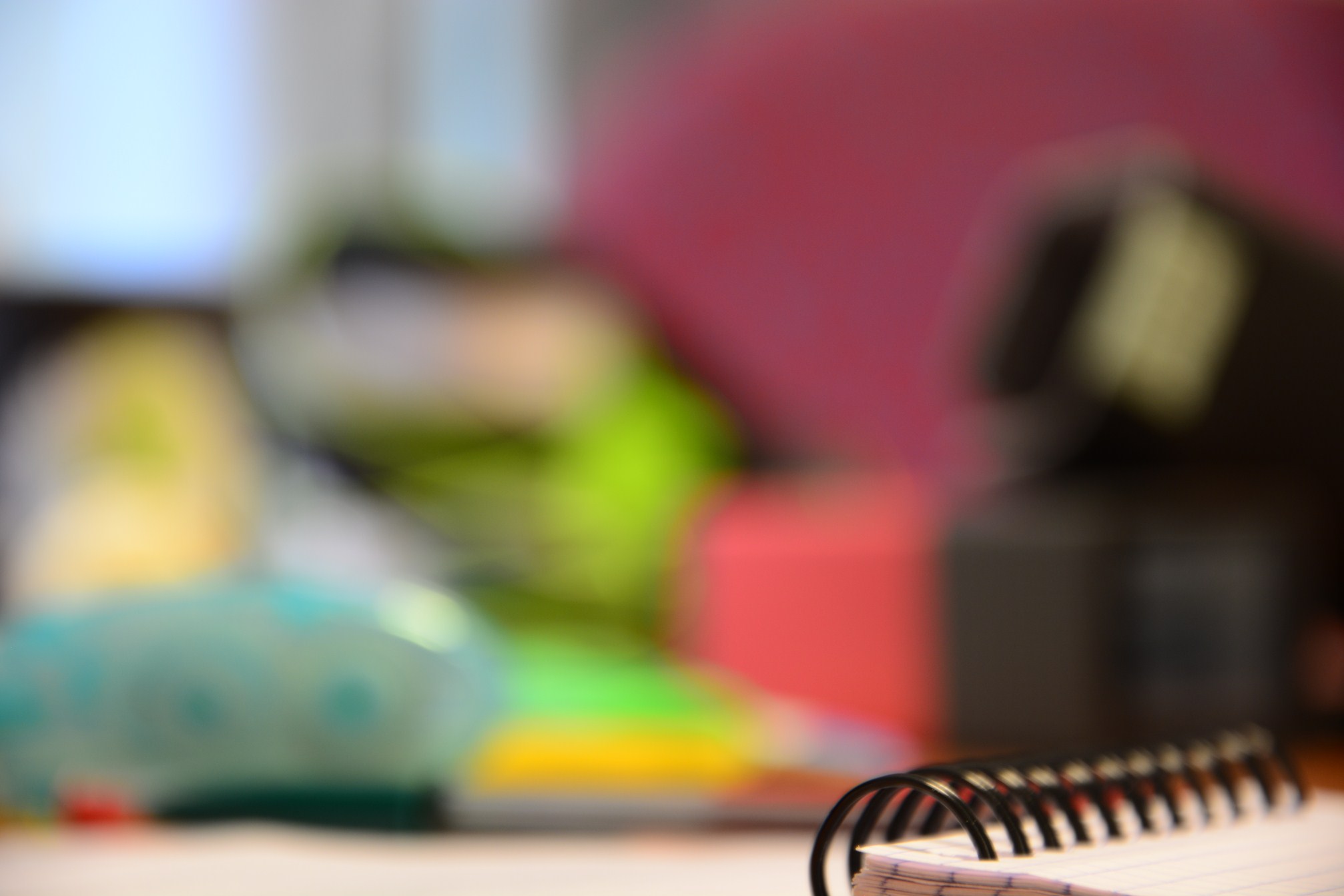}}
\centering\resizebox*{3cm}{!}{\includegraphics{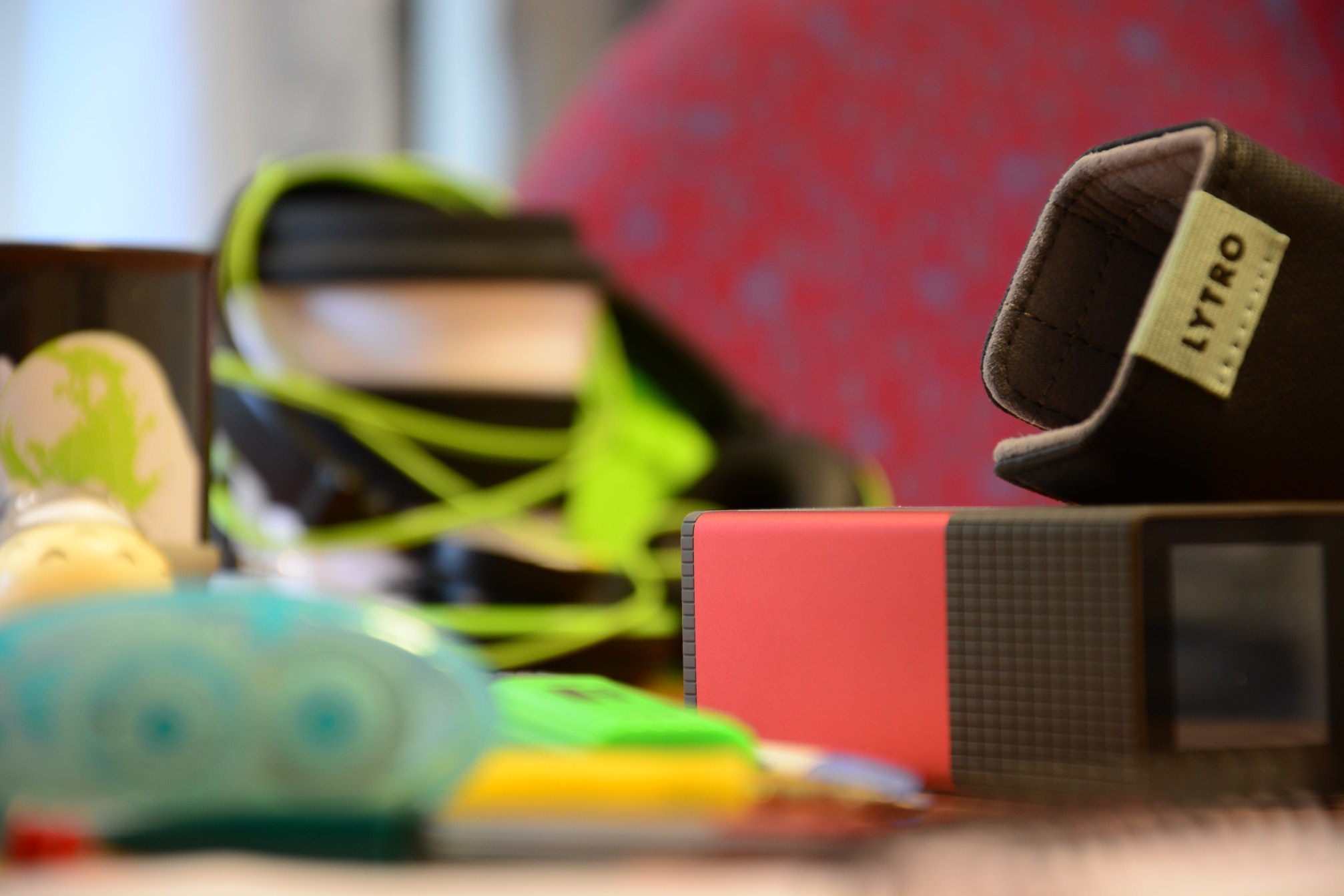}}
\centering\resizebox*{3cm}{!}{\includegraphics{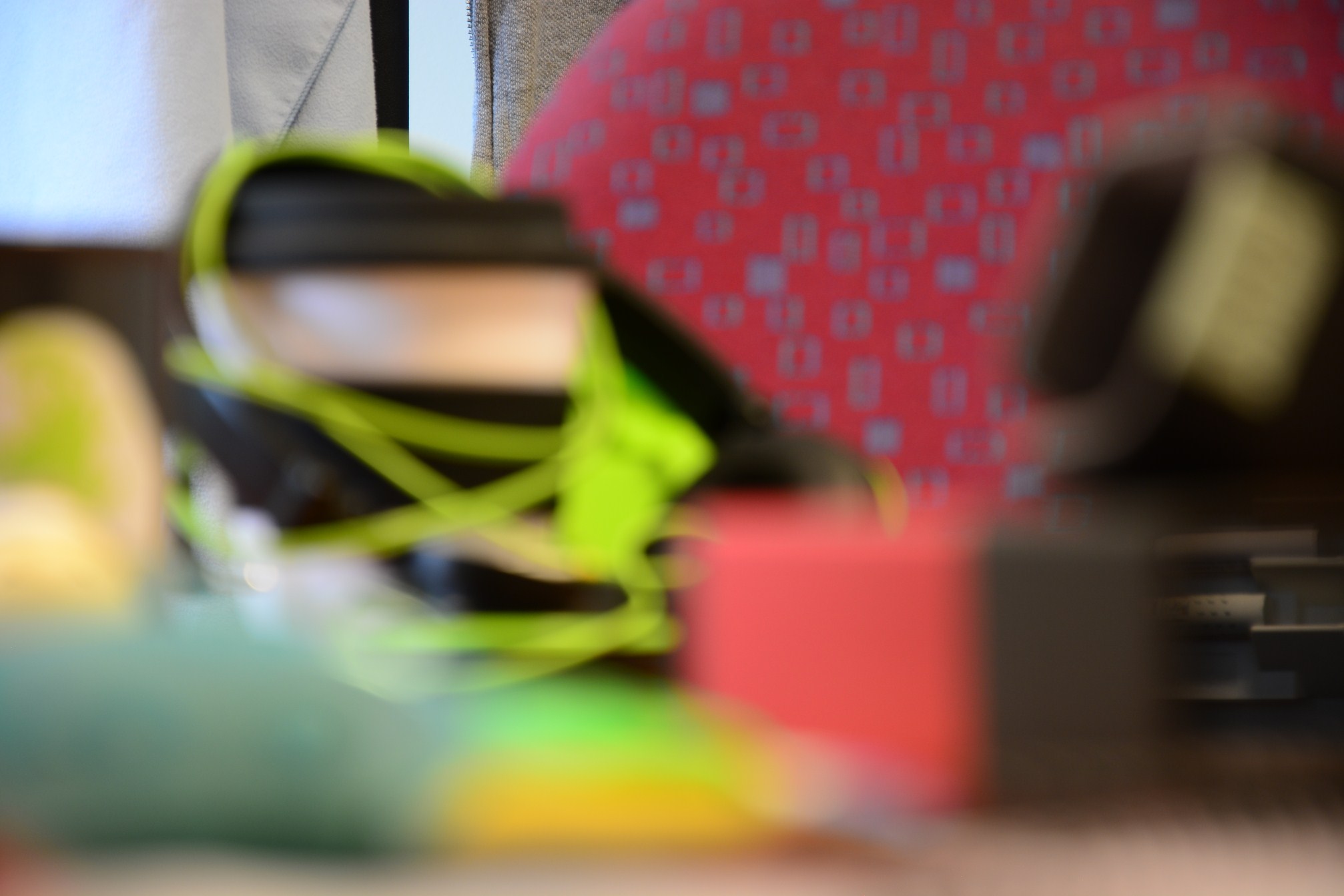}}
\centering\resizebox*{3.2cm}{!}{\includegraphics{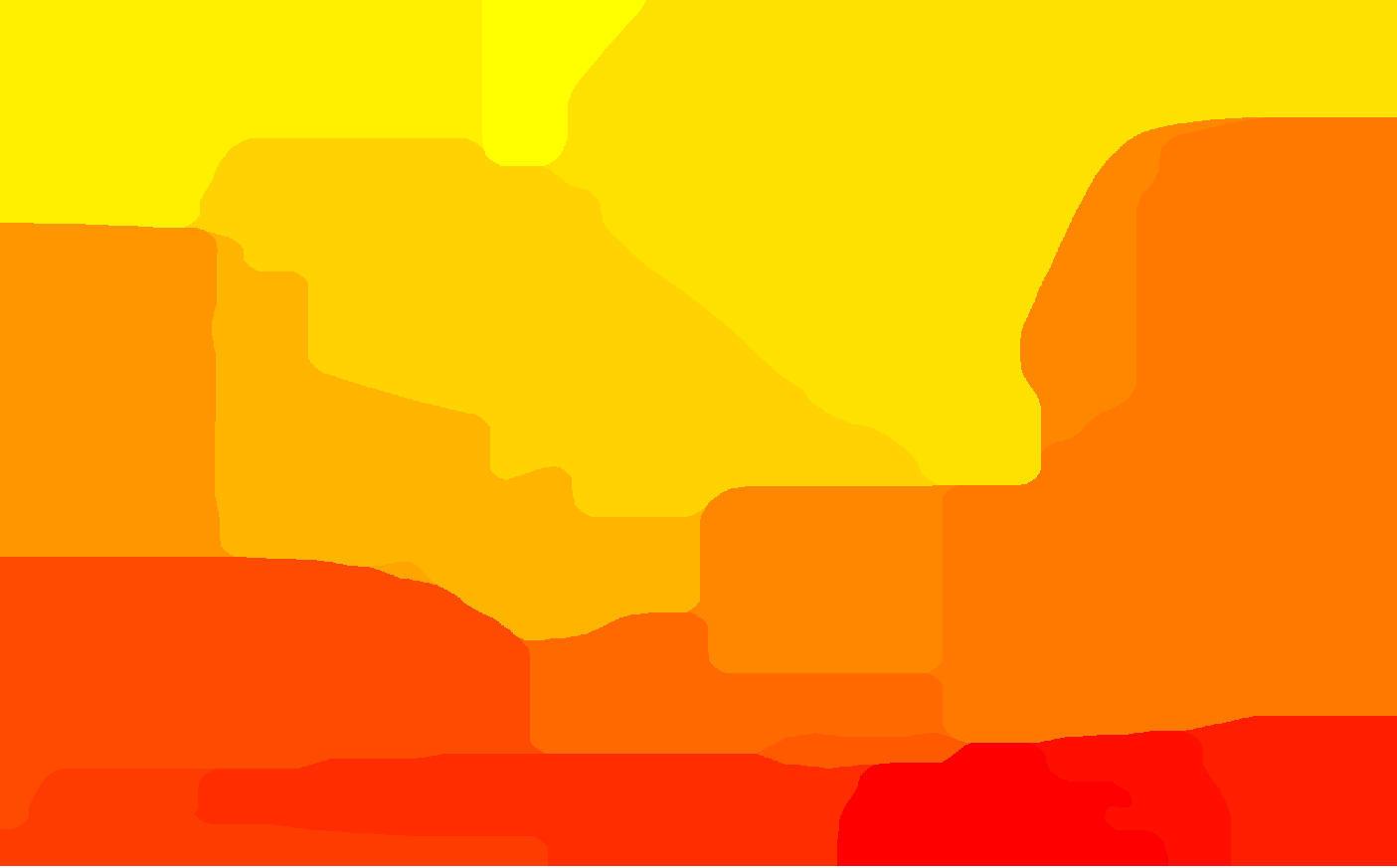}}
\centering\resizebox*{3cm}{!}{\includegraphics{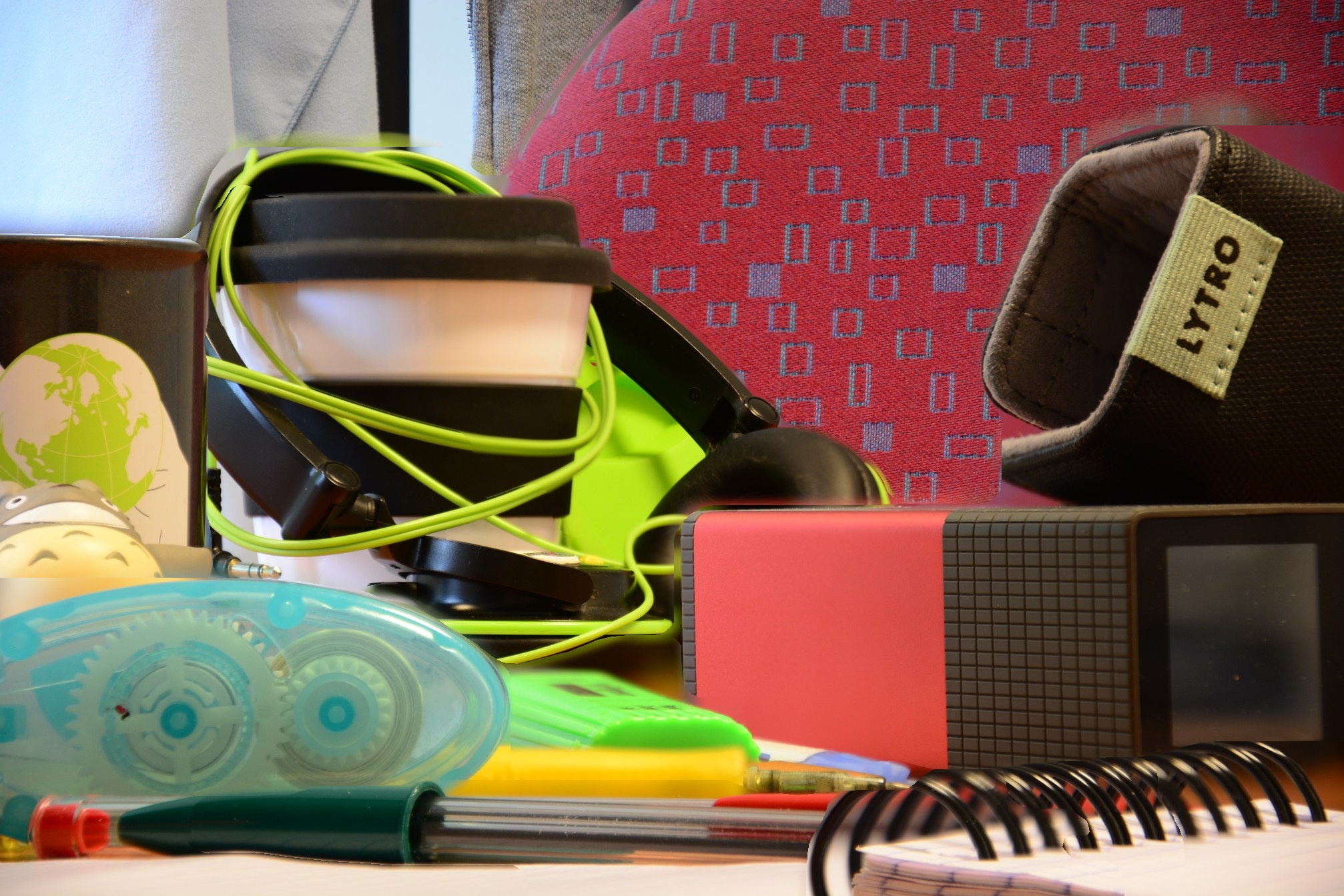}}\\
\vspace{0.1cm}
\centering\resizebox*{3cm}{!}{\includegraphics{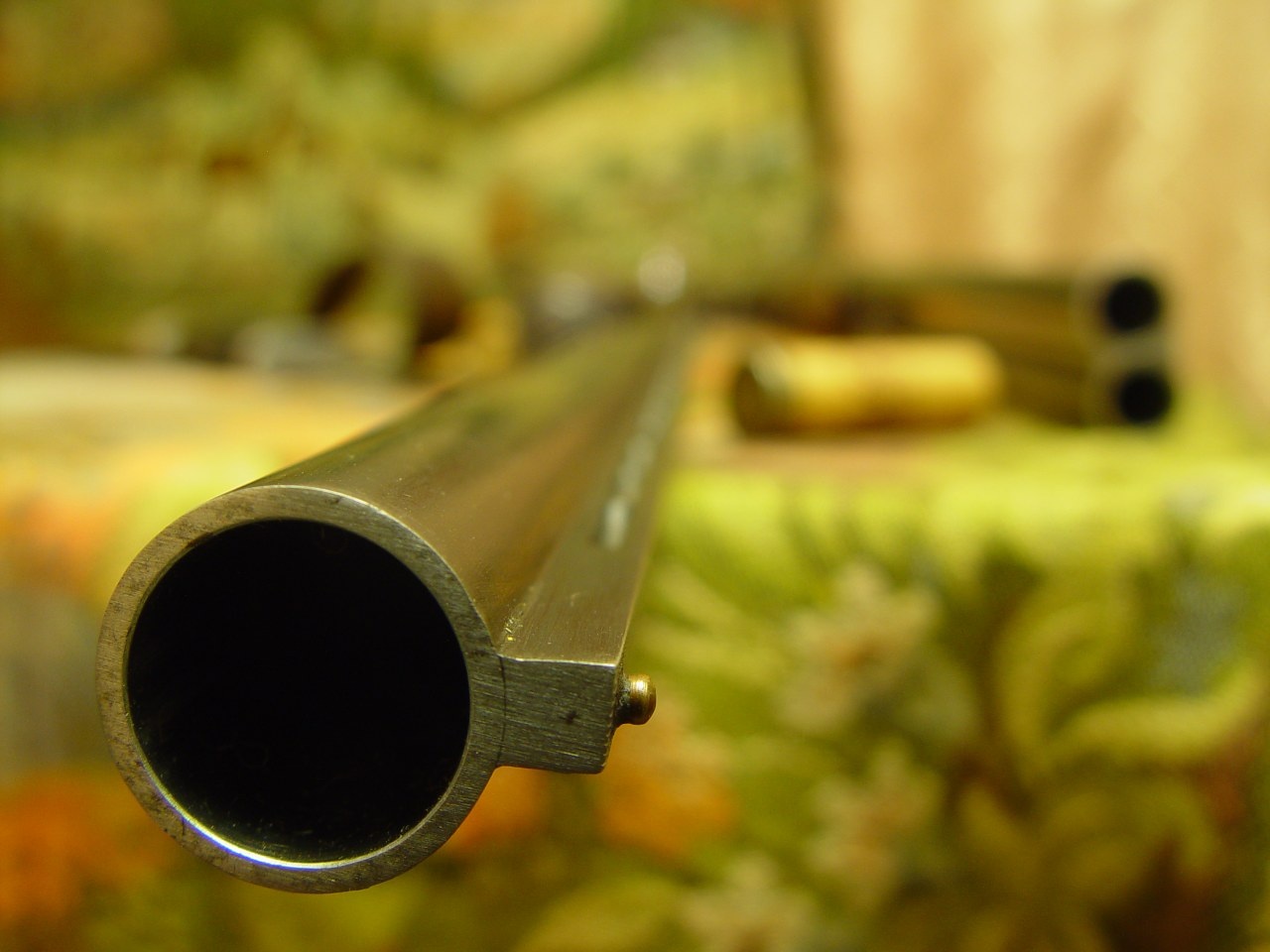}}
\centering\resizebox*{3cm}{!}{\includegraphics{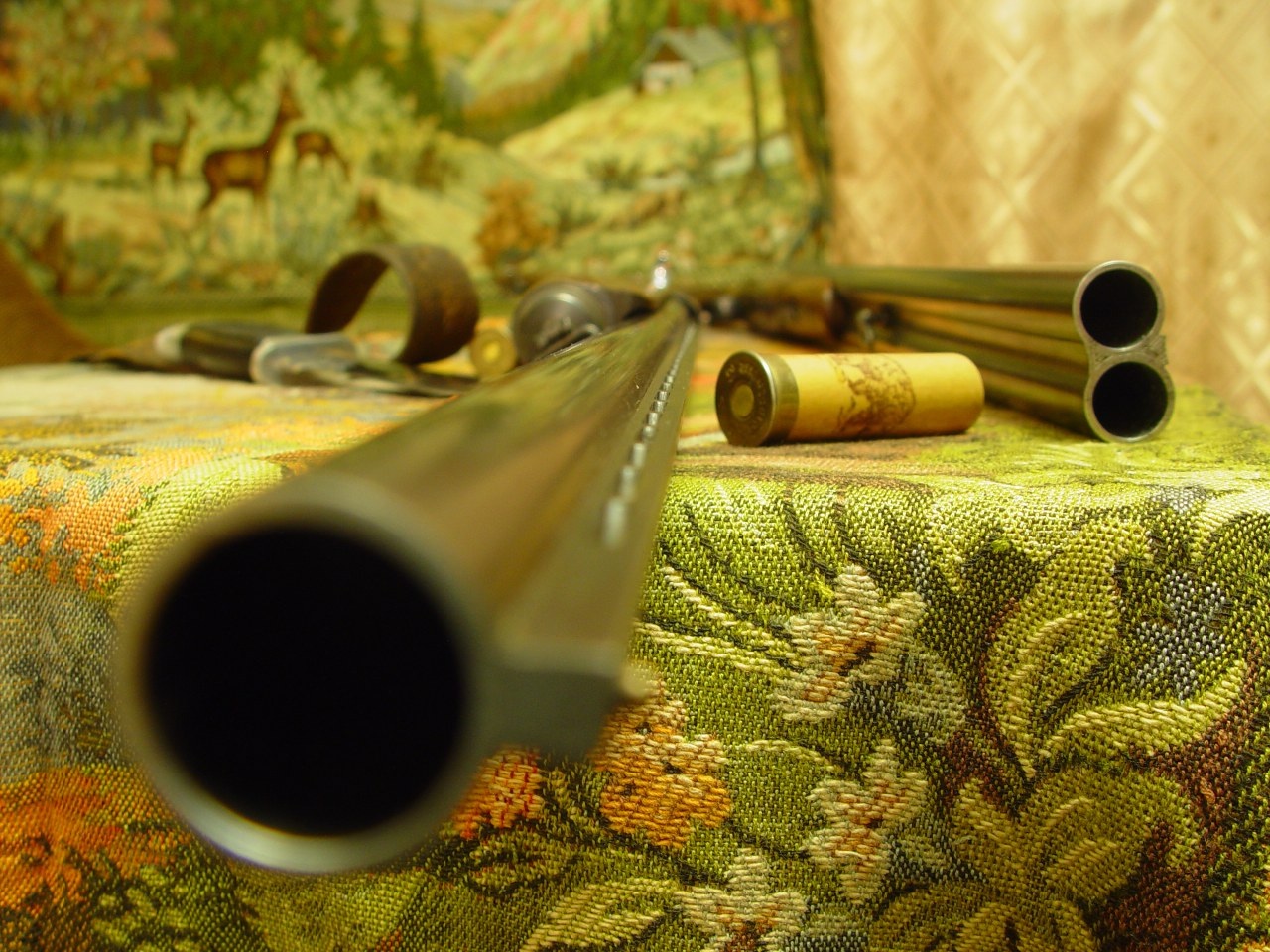}}
\centering\resizebox*{3cm}{!}{\includegraphics{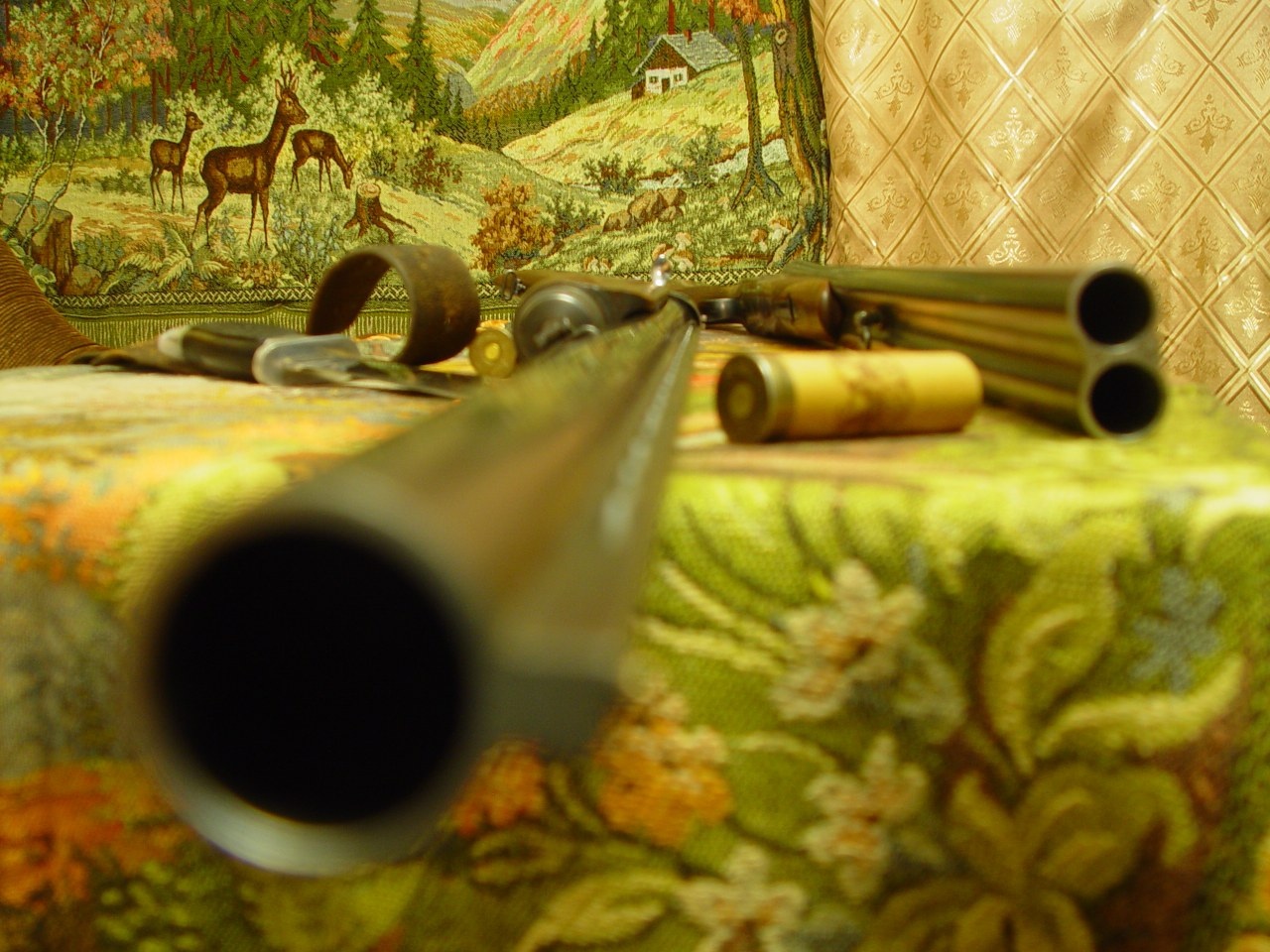}}
\centering\resizebox*{3.1cm}{!}{\includegraphics{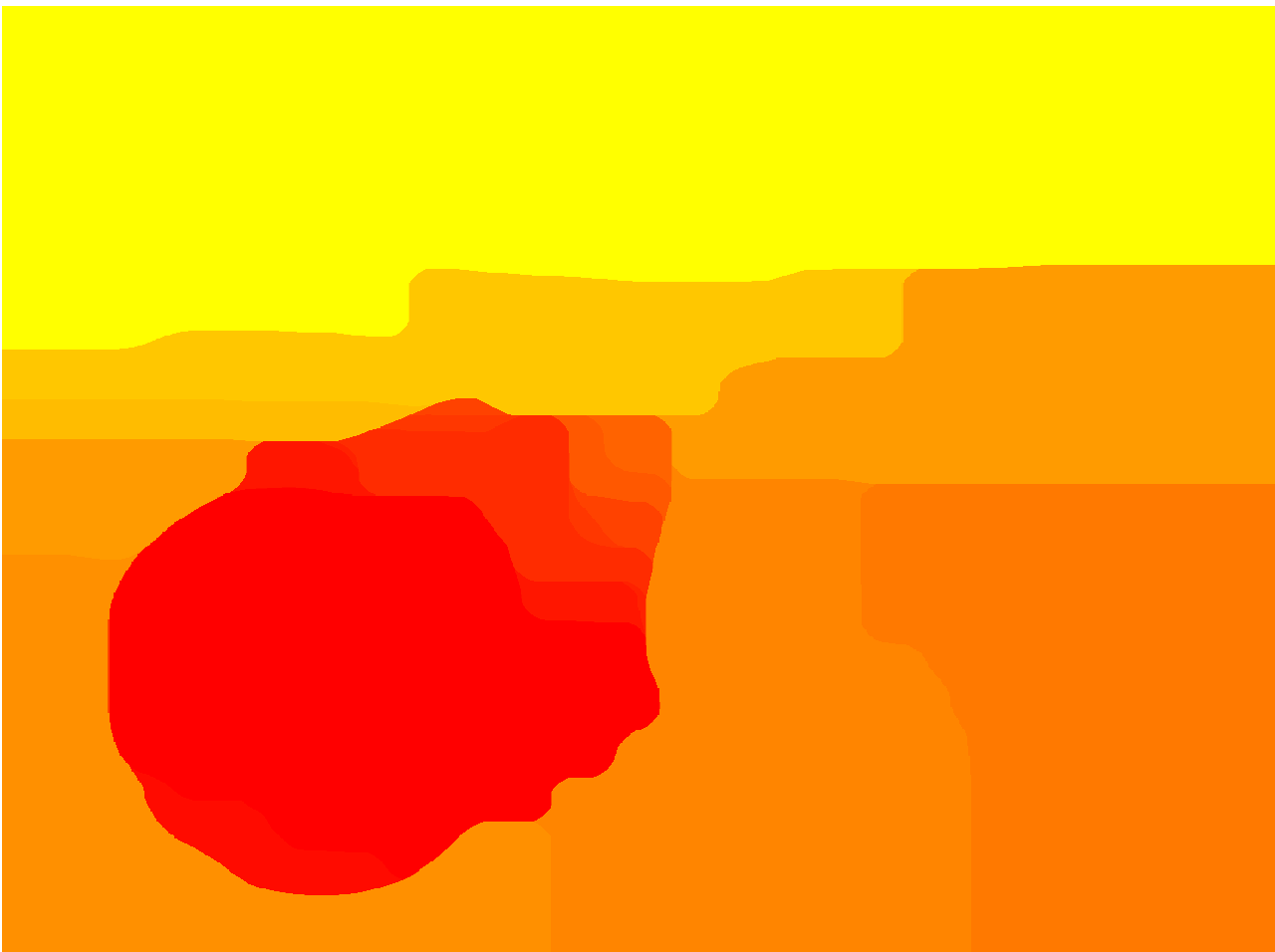}}
\centering\resizebox*{3cm}{!}{\includegraphics{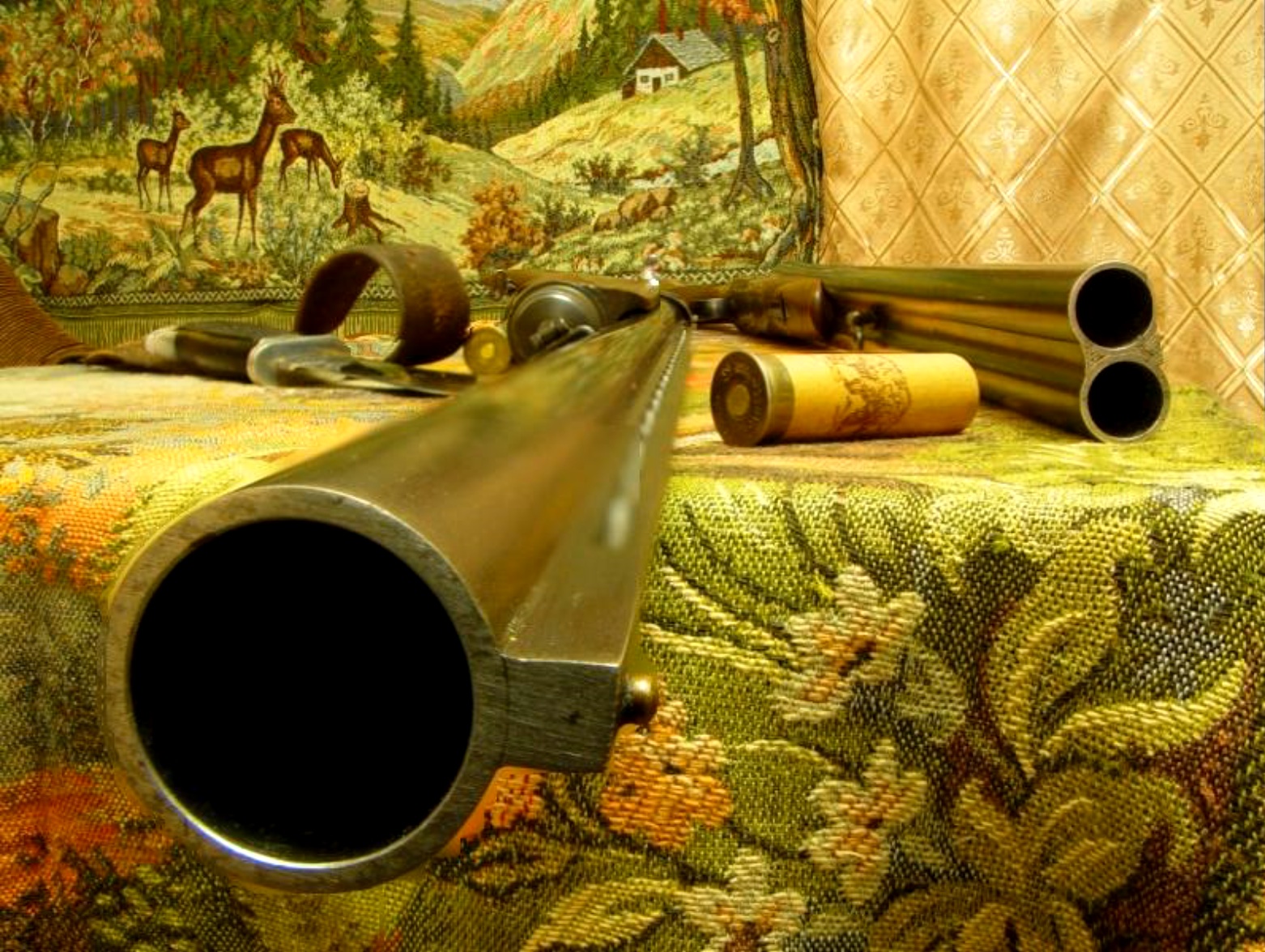}}
\caption{Three images of the focal stack (left); estimated depth map and image with extended focus (right). The focal stack images of the first and second rows have been captured with the Nikon 5200 camera. The focal stack images in the last row come from \cite{Makaruk}.}
\label{dm_ext_Nikon_rifle}
\end{figure*}

\begin{figure*}[tbp]
\vspace{0.1cm}
\centering\resizebox*{2.5cm}{!}{\includegraphics{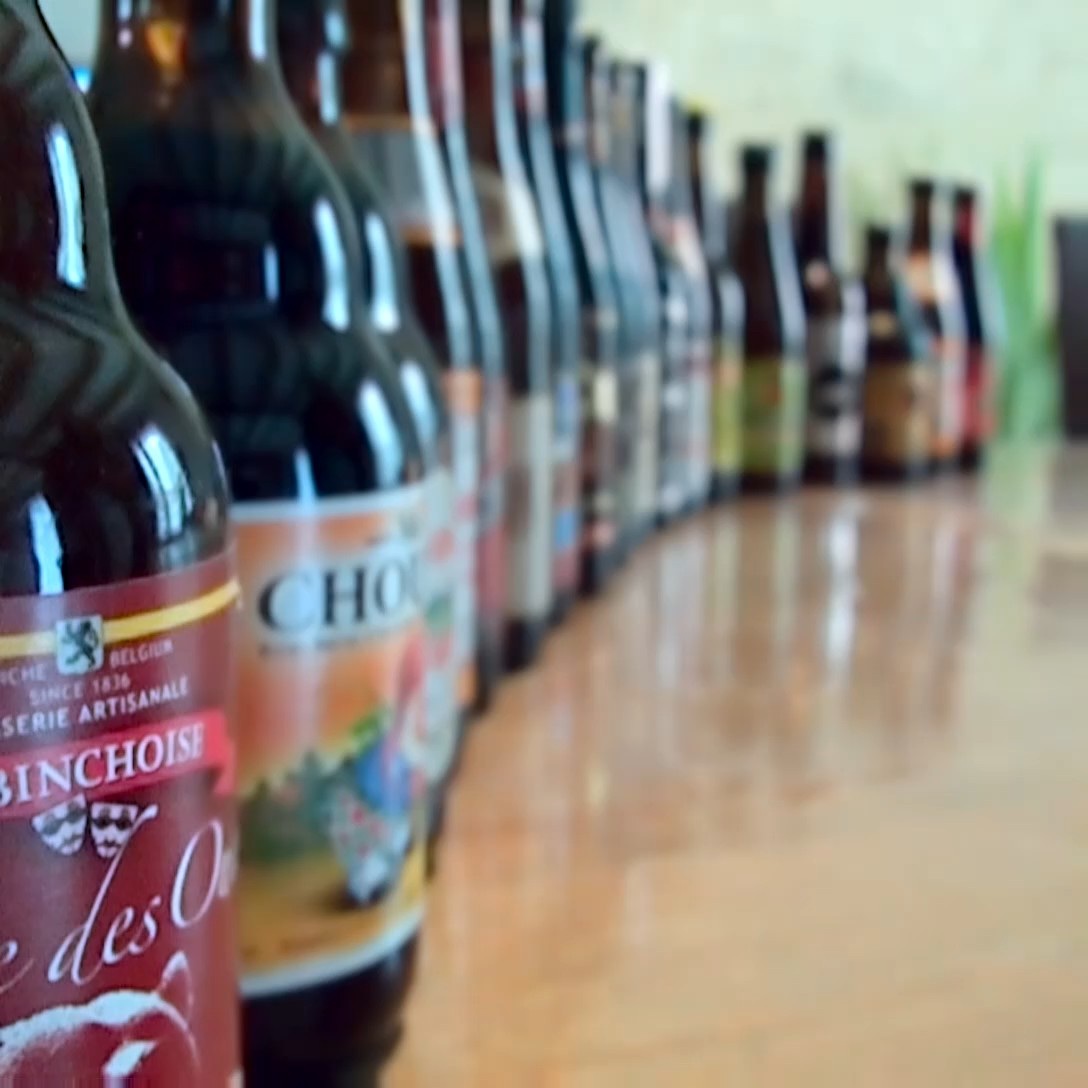}}
\centering\resizebox*{2.5cm}{!}{\includegraphics{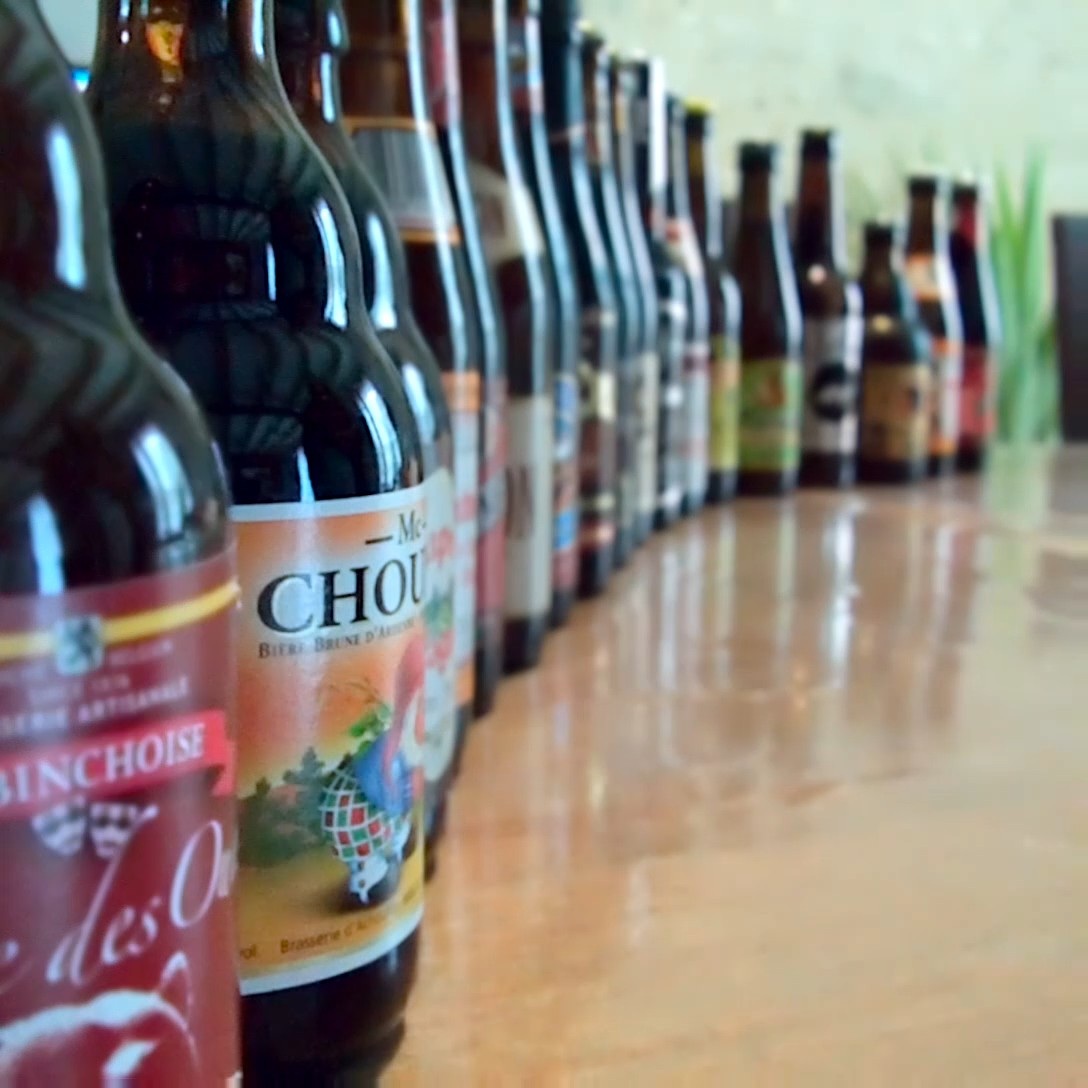}}
\centering\resizebox*{2.5cm}{!}{\includegraphics{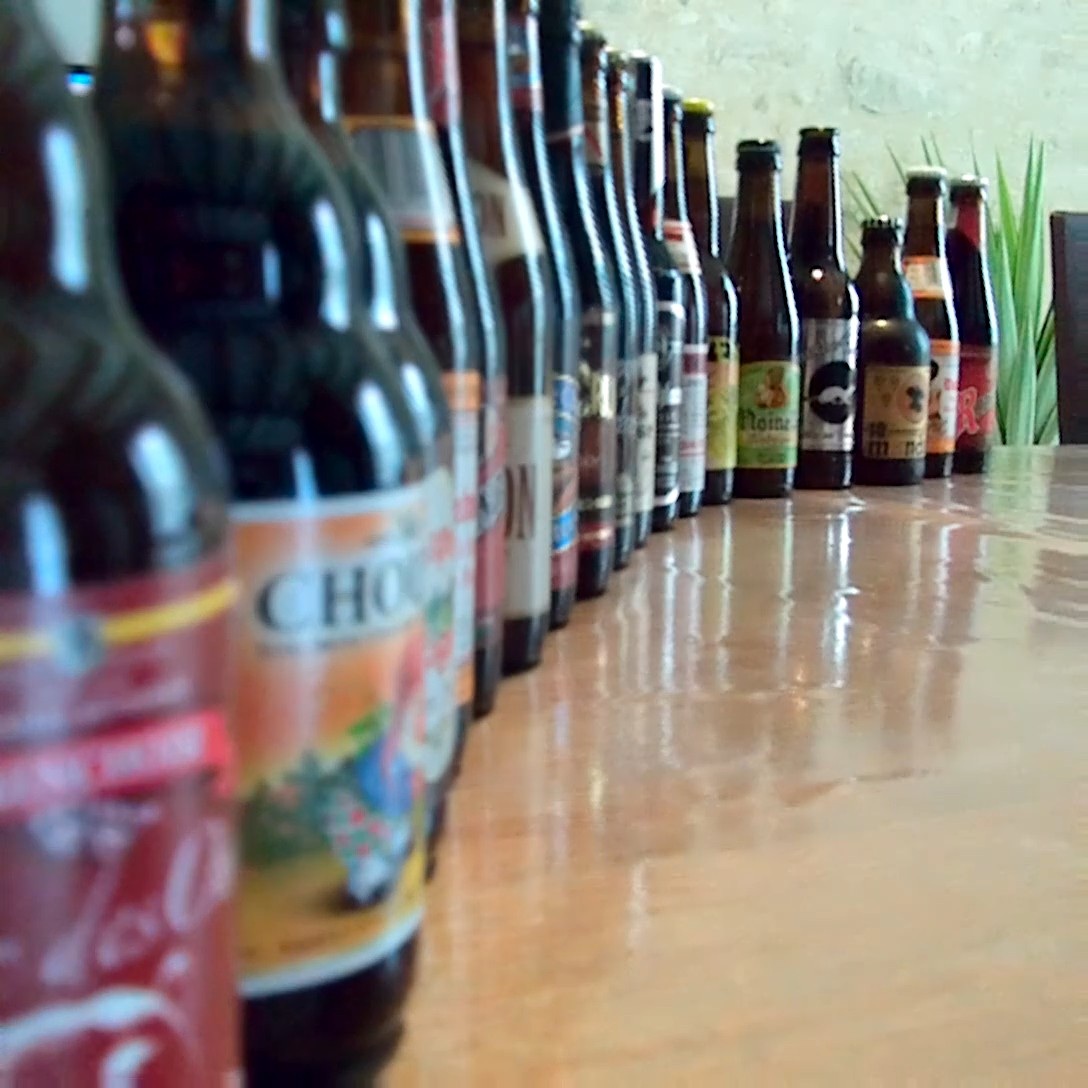}}
\centering\resizebox*{2.54cm}{!}{\includegraphics{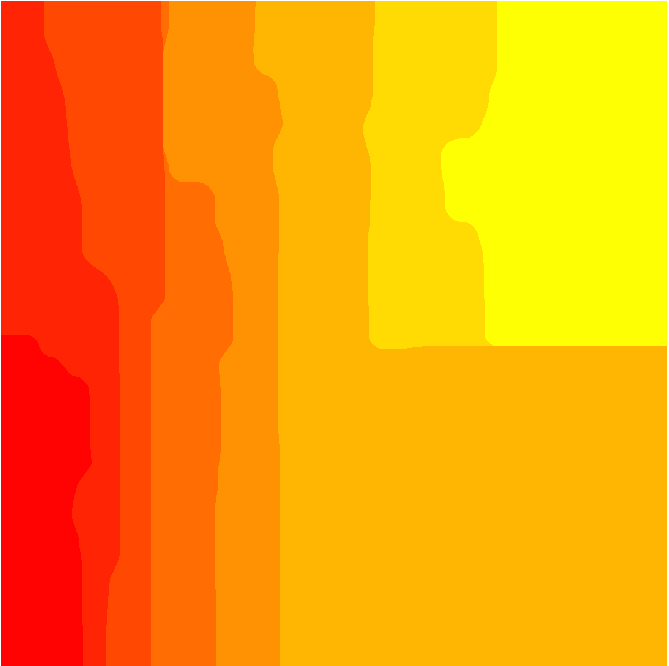}}
\centering\resizebox*{2.5cm}{!}{\includegraphics{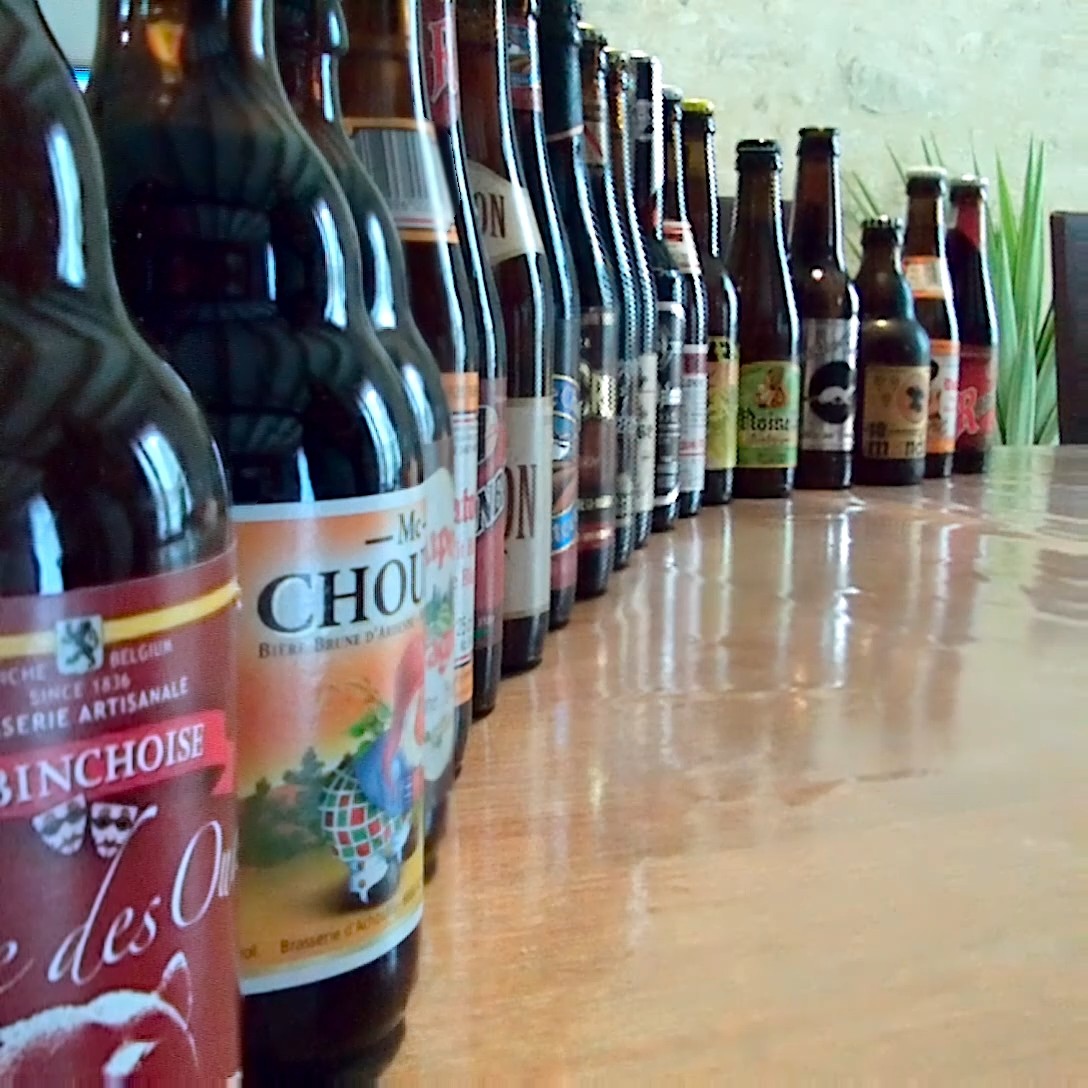}}
\caption{Three images of the focal stack captured by the Lytro camera(left); estimated depth map and image with extended focus (right). }
\label{fig:dm_ext_beers}
\end{figure*}



We evaluate the depth map estimation algorithm proposed in Section \ref{ssec:DMestim} on these three types of data sets. The first two rows of Fig.~\ref{dm_ext_Nikon_rifle} show images captured by the Nikon camera and the last row shows an example taken from \cite{Makaruk}. In 
Fig.~\ref{dm_ext_Nikon_rifle}, the first, middle and last images of the focal stack are shown, as well as the depth maps estimated by gradient thresholding and region growing via graph cut. The depth maps are post-processed with a median filter to correct small artifacts produced by the region growing algorithm.

For the rifle image (third row of Fig.~\ref{dm_ext_Nikon_rifle}), since the calibration information and the metadata of the capture are not available for this data set, we only estimate and present the focus map instead of the depth map. Although a numerical evaluation of the results is not possible due to the lack of ground truth data, the visual inspection of the results suggests that the obtained depth and focus maps are of satisfactory quality. An obvious limitation of the method is the quantization of the depth values as dictated by the number of images in the focal stack. Nevertheless, the image synthesis experiments in Sections \ref{ssec:exp_ext_focus} and \ref{ssec:exp_pers_shift}, which use the depth maps estimated with the proposed approach, show that it is possible to obtain a quite good image rendering quality in spite of the quantization of the depth map.

\subsection{Extended focus}
\label{ssec:exp_ext_focus}

The estimated focus maps can be used to render an all-in-focus image, which is called extended focus \cite{Ng}. The all-in-focus image is formed by merging the in-focus regions of the focal stack images. 
The rightmost column of Fig.~\ref{dm_ext_Nikon_rifle} shows the extended focus images generated with the help of the focus map. The rendered image is of satisfactory quality, while some artifacts can be observed on the border between the card and the shoe for the first image. They are due to focus map estimation errors caused by the fact that the card is textureless near the boundary, while the shoe contains more texture and thus has stronger image intensity gradients. This results in a slight underestimation of the region where the card is in focus.
Fig.~\ref{fig:dm_ext_beers} shows some images from the focal stack of a scene captured with the Lytro camera, the estimate of the focus map with the proposed algorithm, and the extended focus image computed with the proposed algorithm. Some imprecision in the focus map is observable in the part of the image corresponding to the table, which is a problematic non-Lambertian surface lacking texture. However, it is seen that these errors do not affect much the quality of the extended focus image. The results show that the presented method can be successfully used for rendering an image that is in-focus everywhere. 


\begin{figure*}[th]
\centering\resizebox*{5cm}{!}{\includegraphics{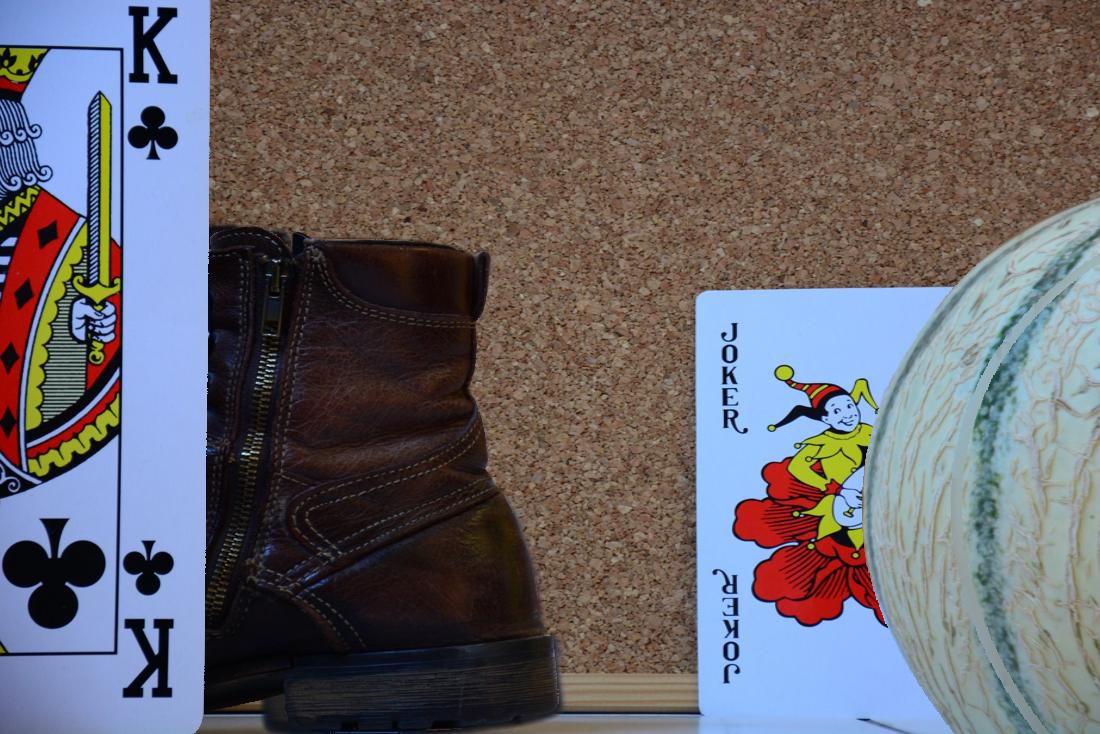}}
\centering\resizebox*{5cm}{!}{\includegraphics{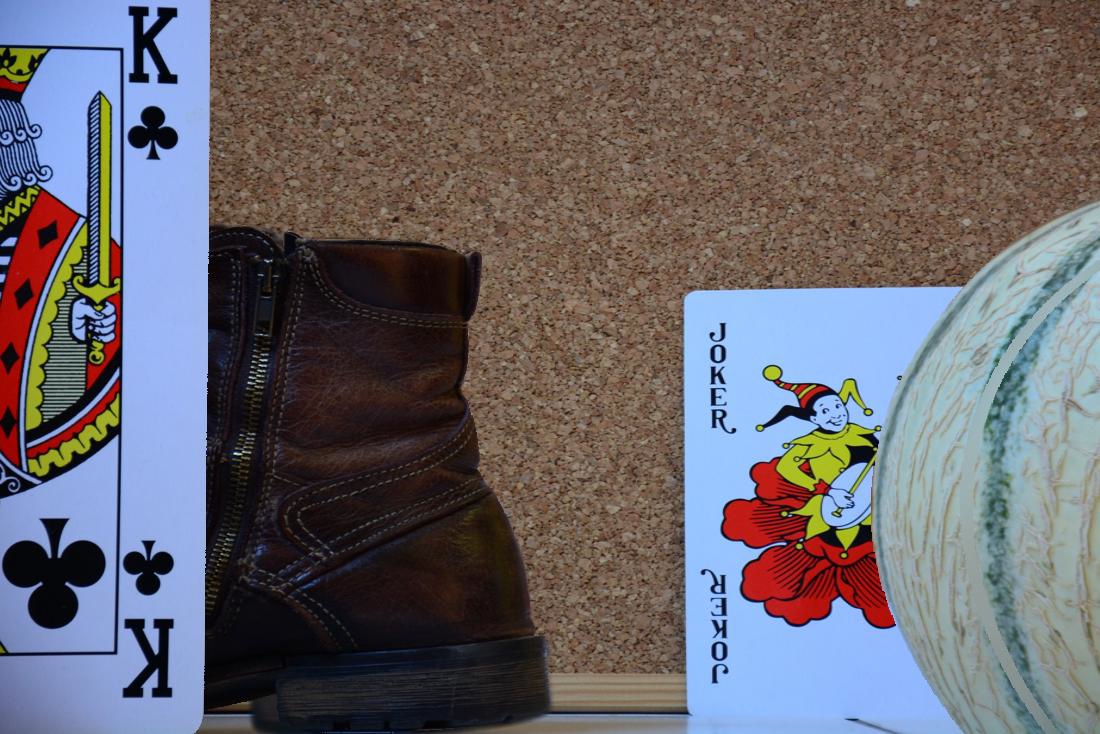}}
\centering\resizebox*{5cm}{!}{\includegraphics{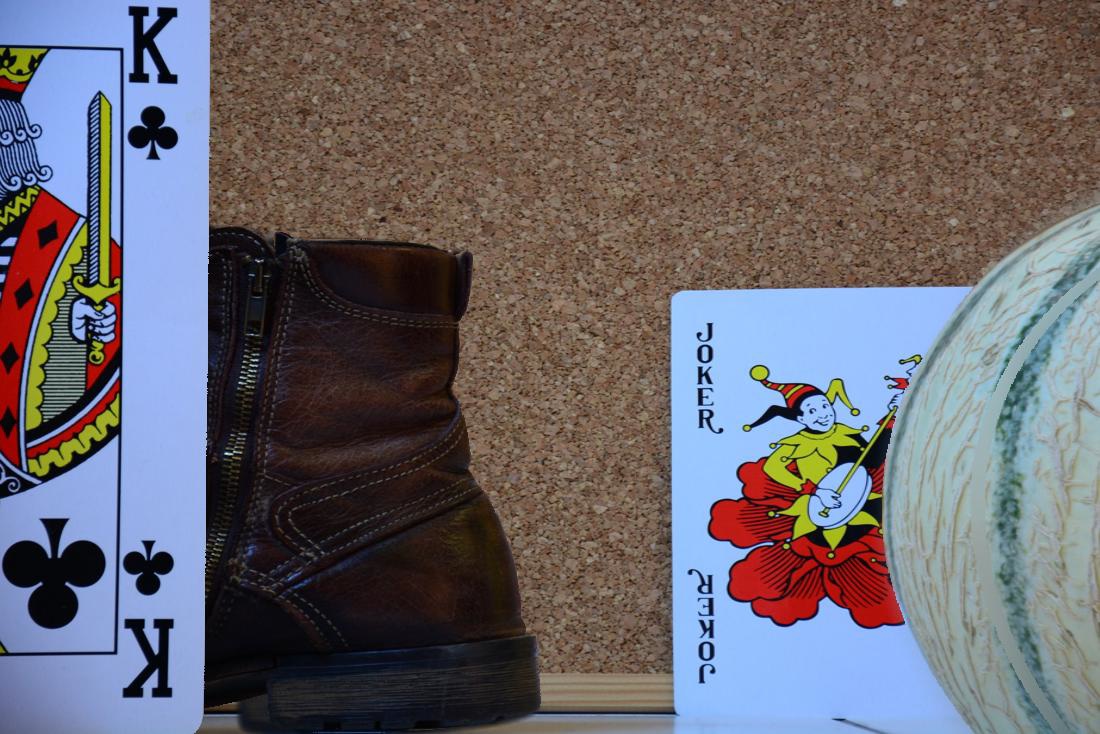}}\\
\vspace{0.1cm}
\centering\resizebox*{5cm}{!}{\includegraphics{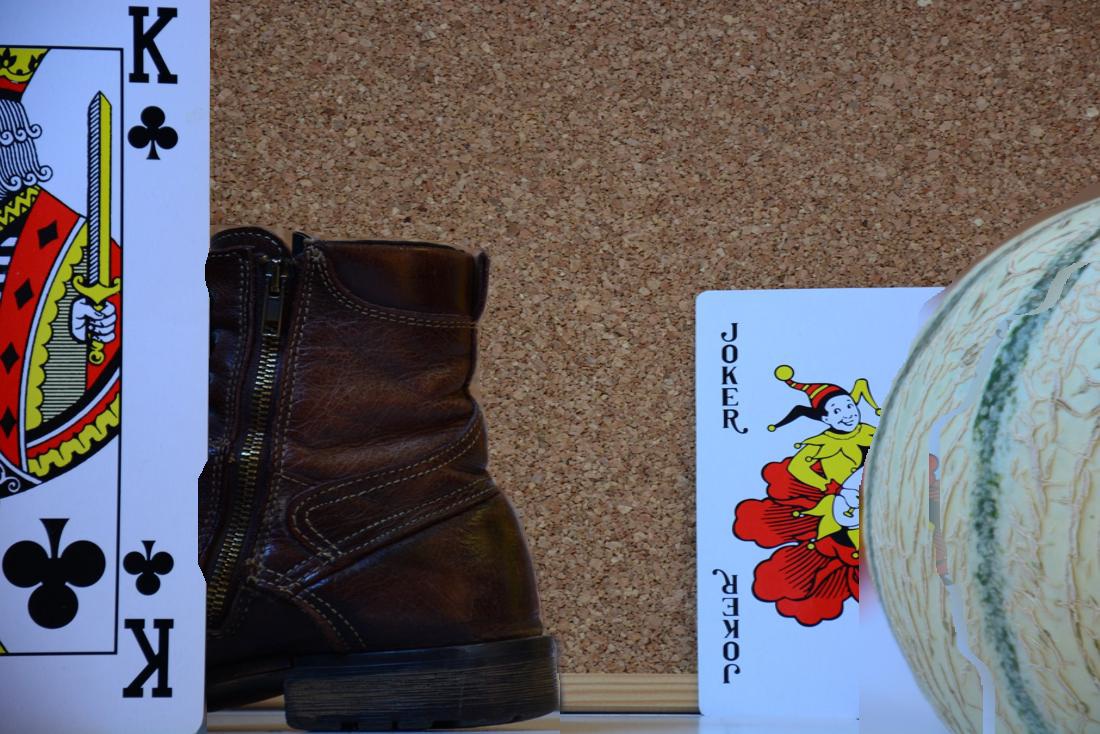}}
\centering\resizebox*{5cm}{!}{\includegraphics{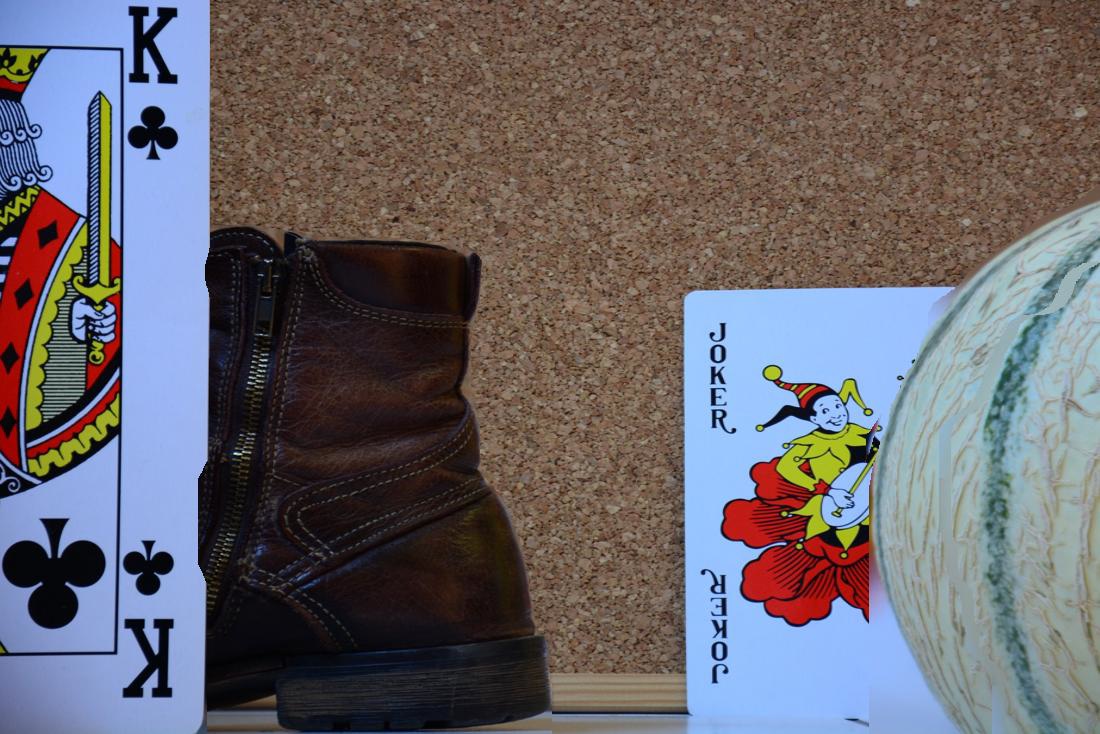}}
\centering\resizebox*{5cm}{!}{\includegraphics{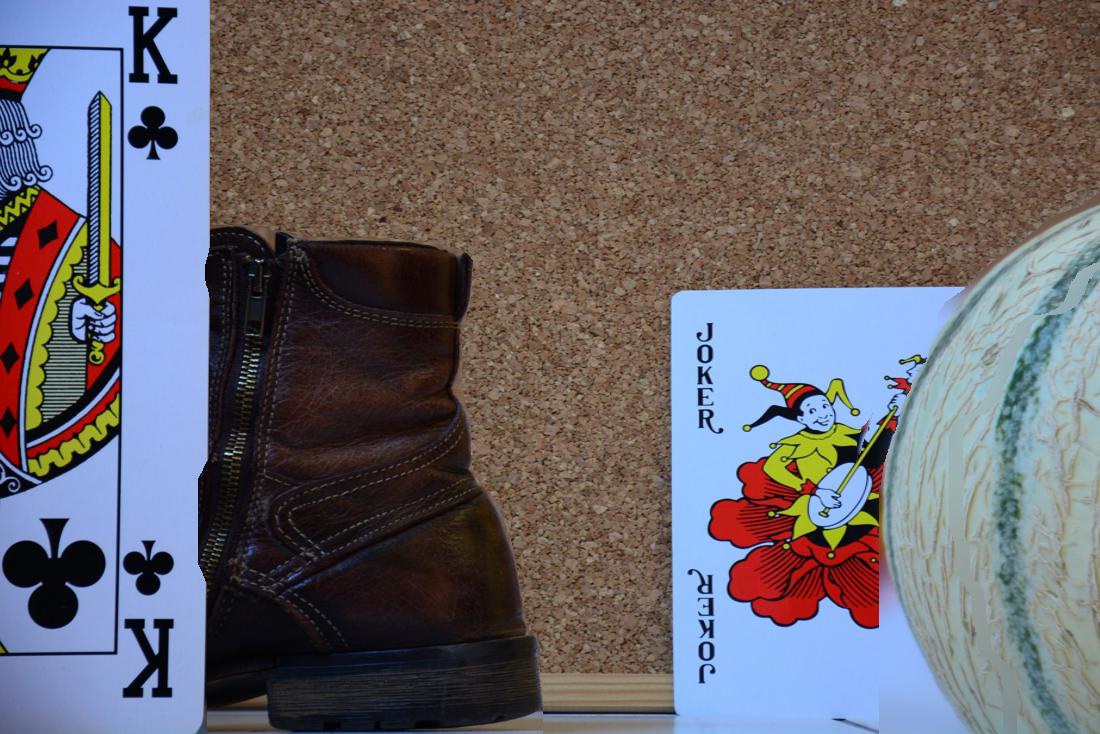}}\\
\vspace{0.1cm}
\centering\resizebox*{5cm}{!}{\includegraphics{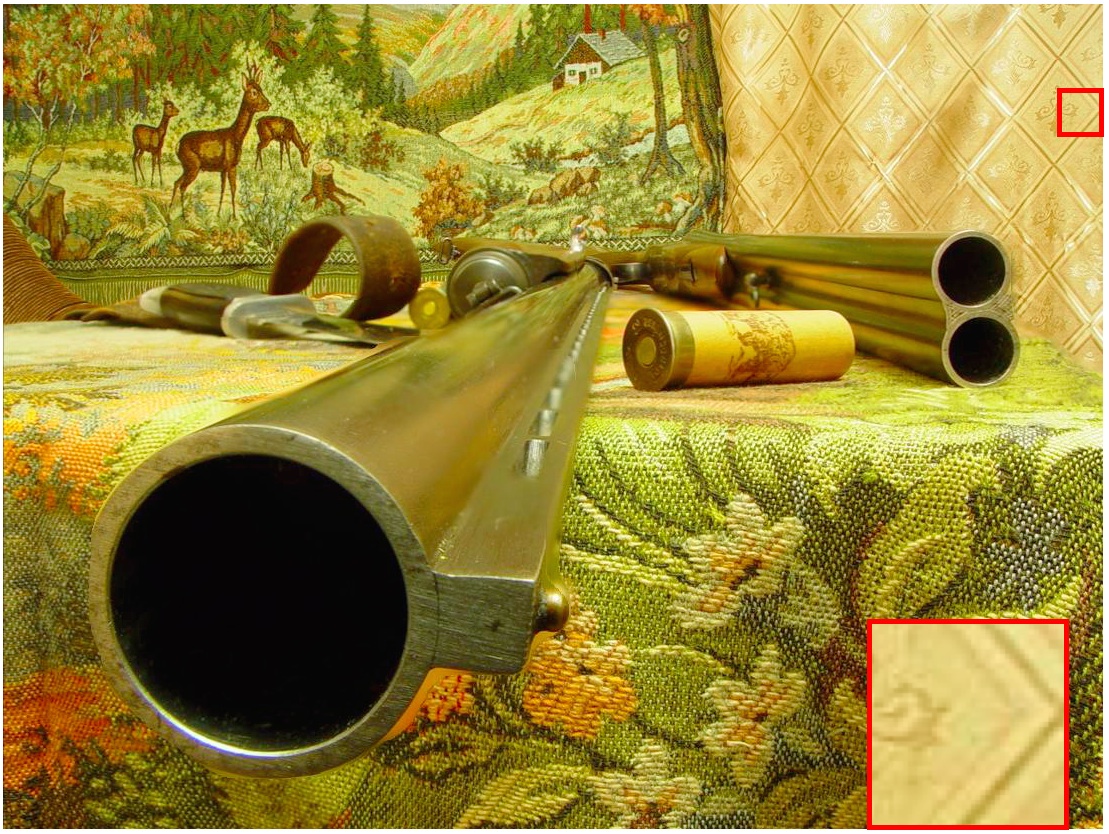}}
\centering\resizebox*{5cm}{!}{\includegraphics{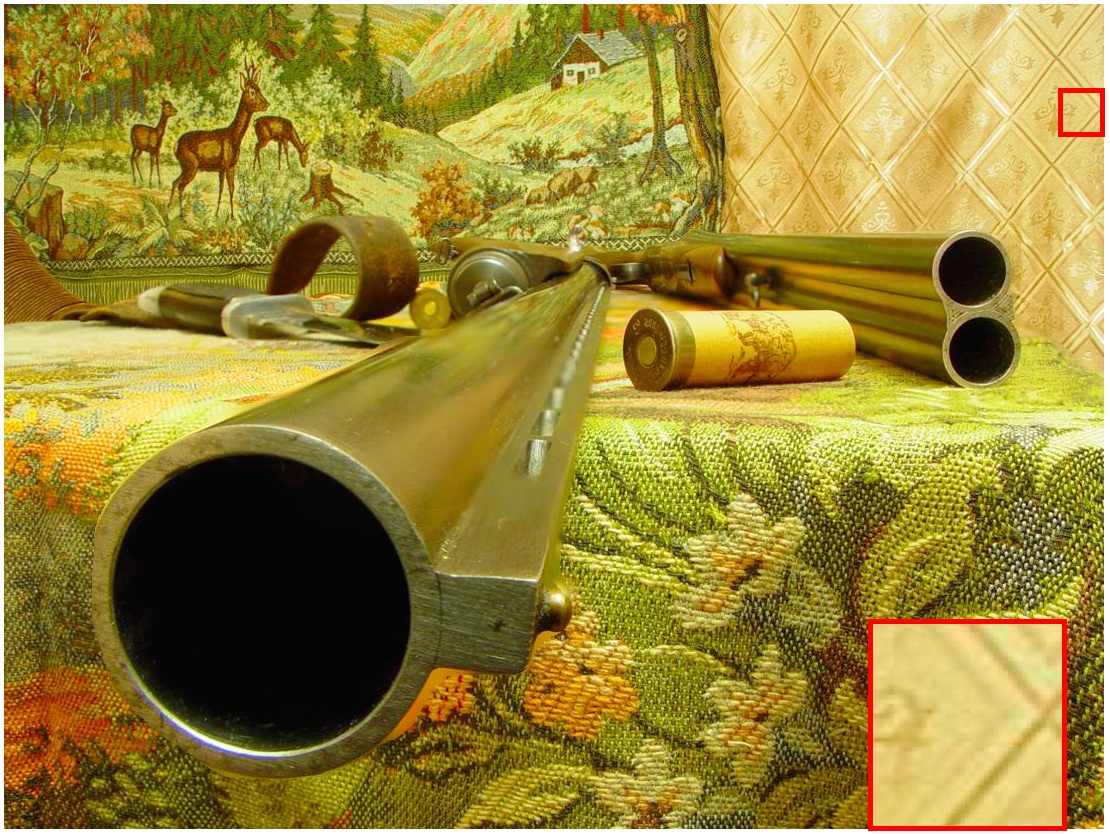}}
\centering\resizebox*{5cm}{!}{\includegraphics{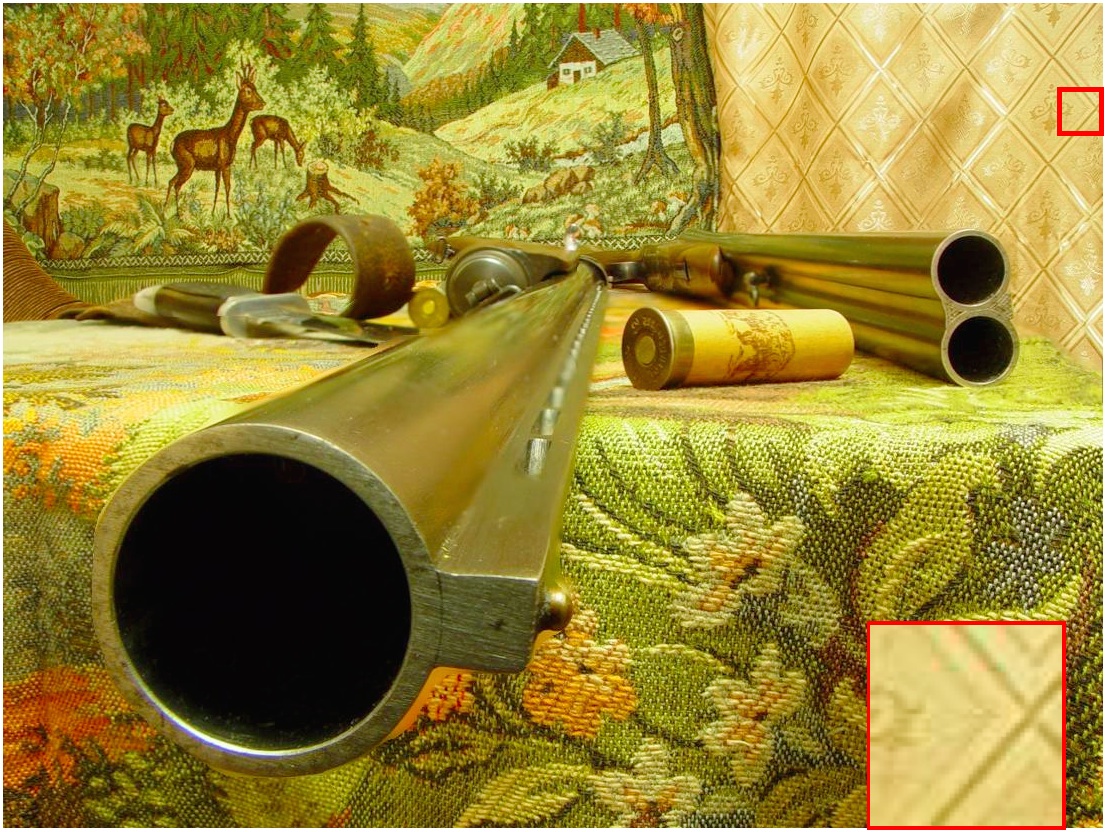}}\\
\caption{First, middle, and last frames of the perspective shifts rendered with the proposed method. In the upper two rows, the left hand of the joker is not visible in the first frame, whereas it becomes visible in the last one. In the bottom row, the viewpoint change is noticeable along the left and right borders of the image (a close-up of the region selected along the right border is shown)}
\label{persp}
\vspace{-10pt}
\end{figure*}

\begin{figure*}[th]
\centering\resizebox*{5cm}{!}{\includegraphics{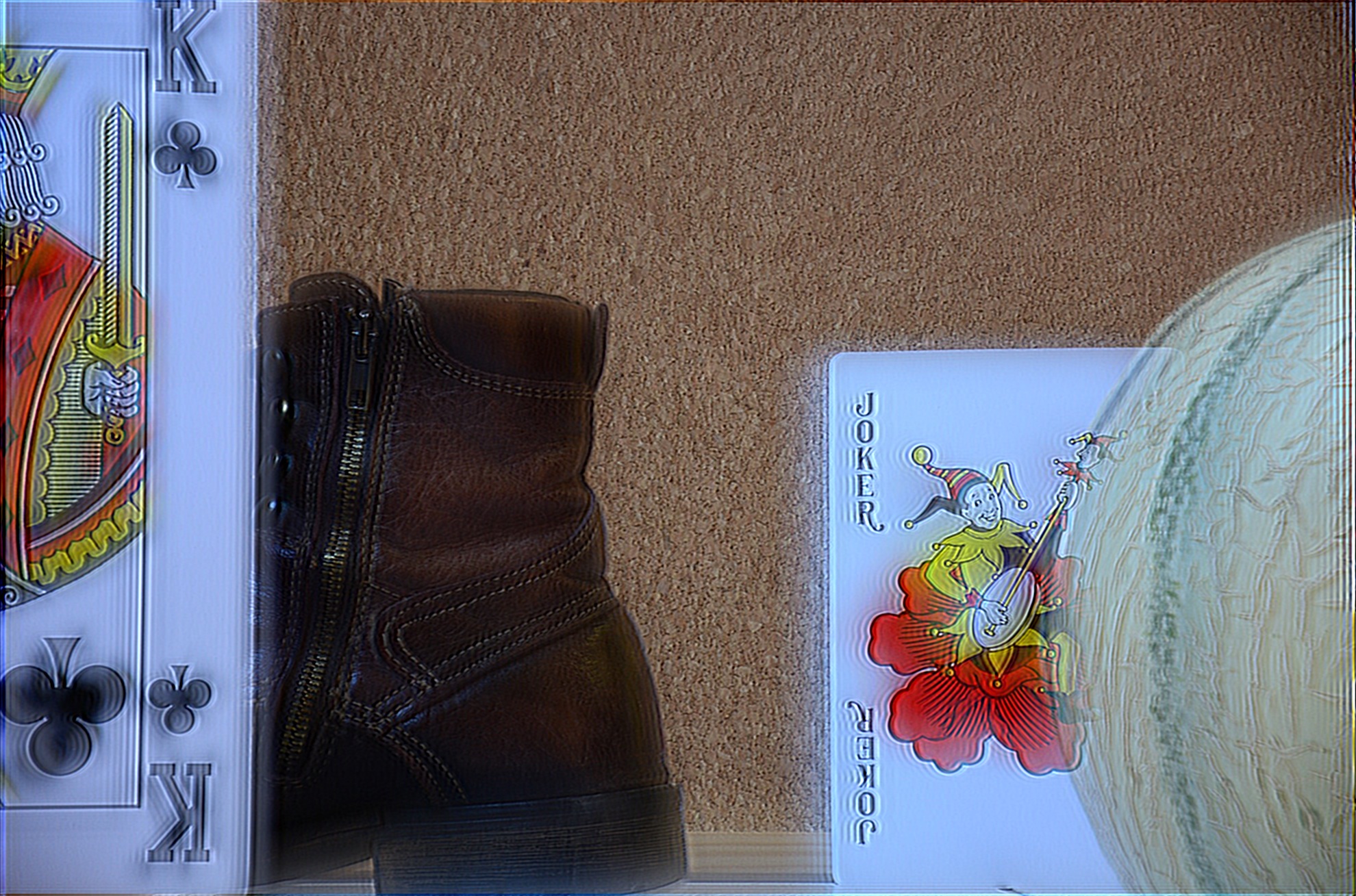}}
\centering\resizebox*{5cm}{!}{\includegraphics{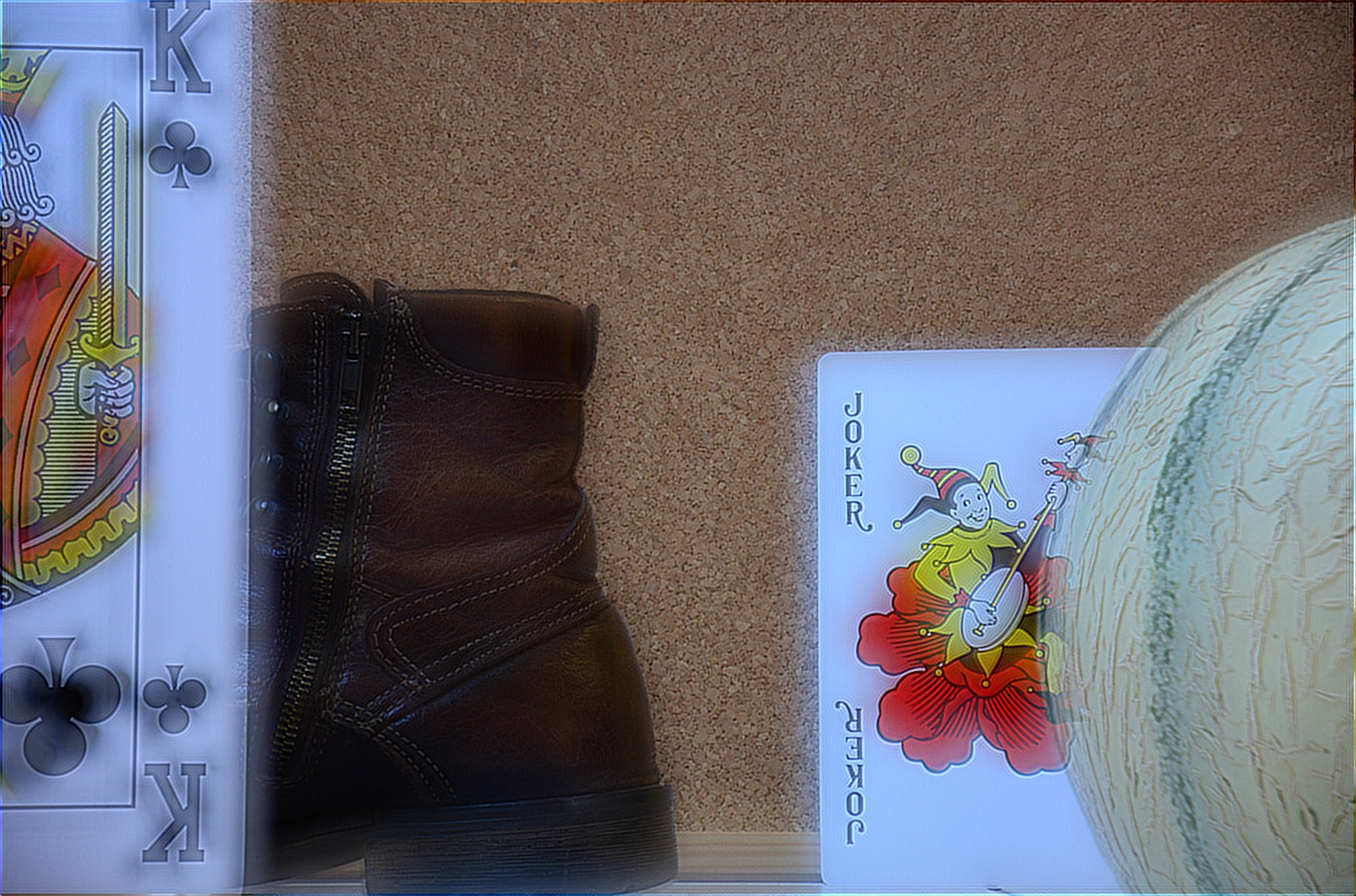}}
\centering\resizebox*{5cm}{!}{\includegraphics{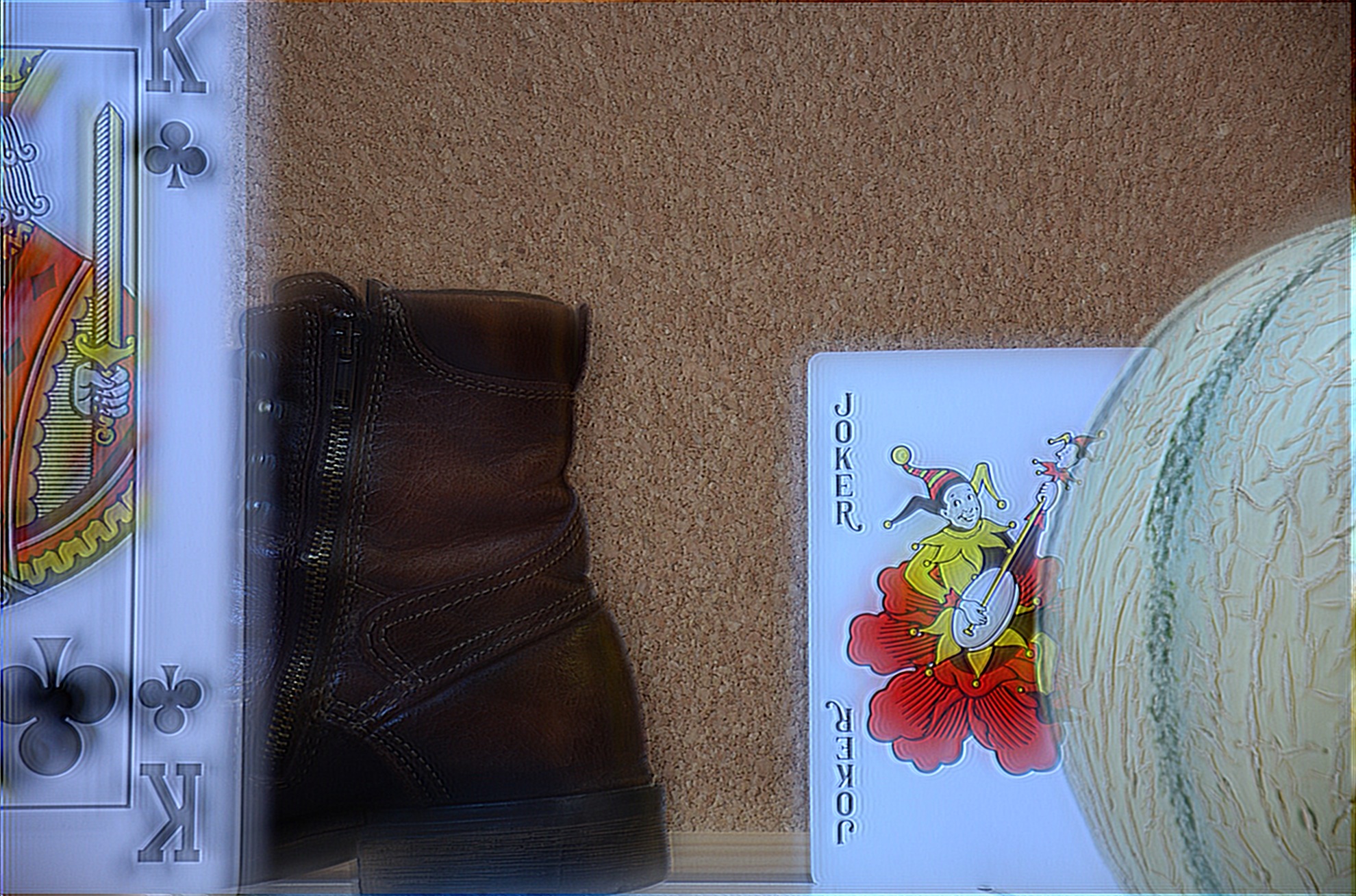}}\\
\vspace{0.1cm}
\centering\resizebox*{5cm}{!}{\includegraphics{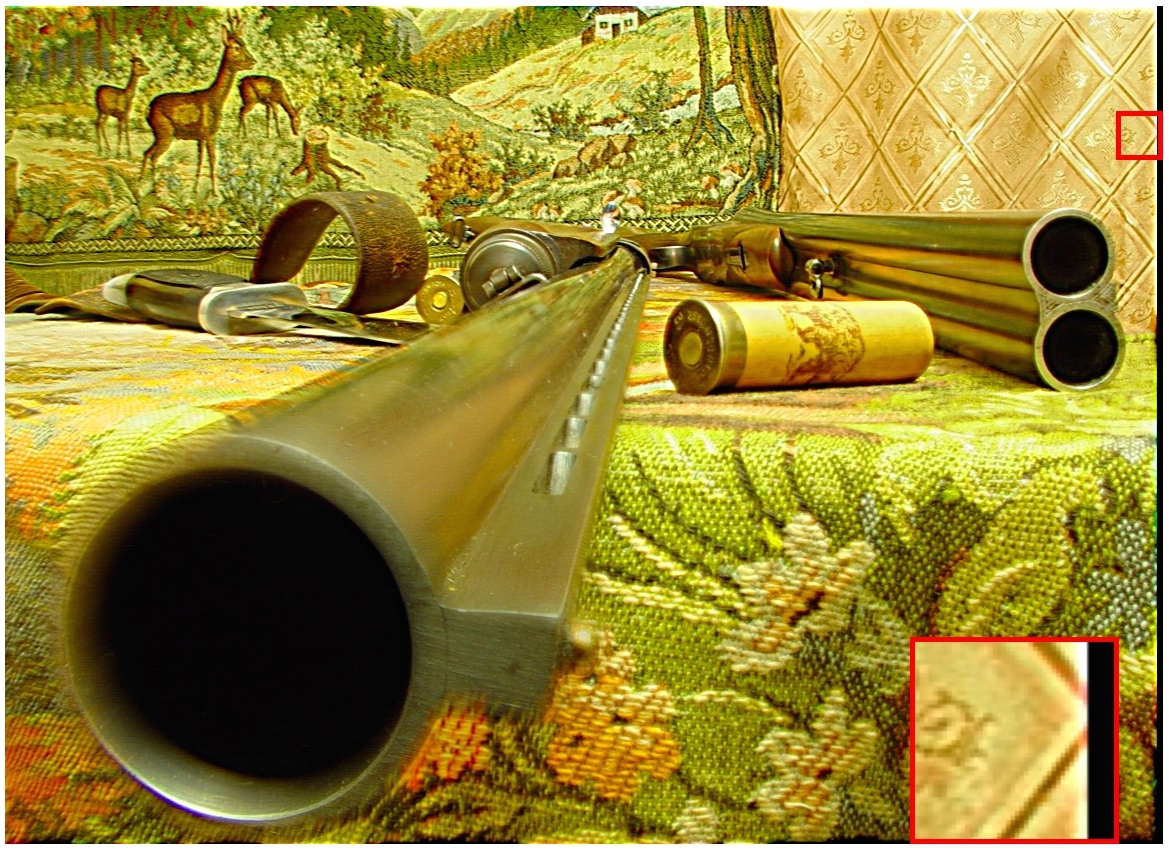}}
\centering\resizebox*{5cm}{!}{\includegraphics{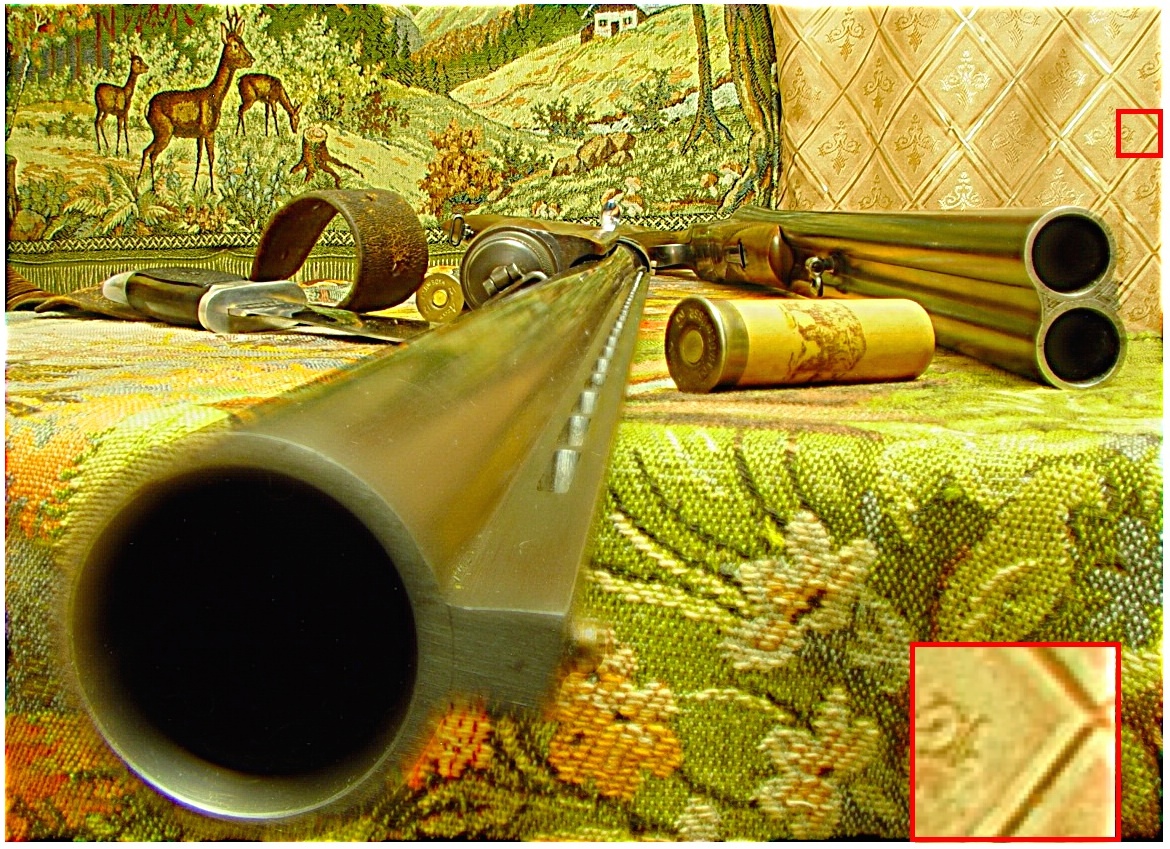}}
\centering\resizebox*{5cm}{!}{\includegraphics{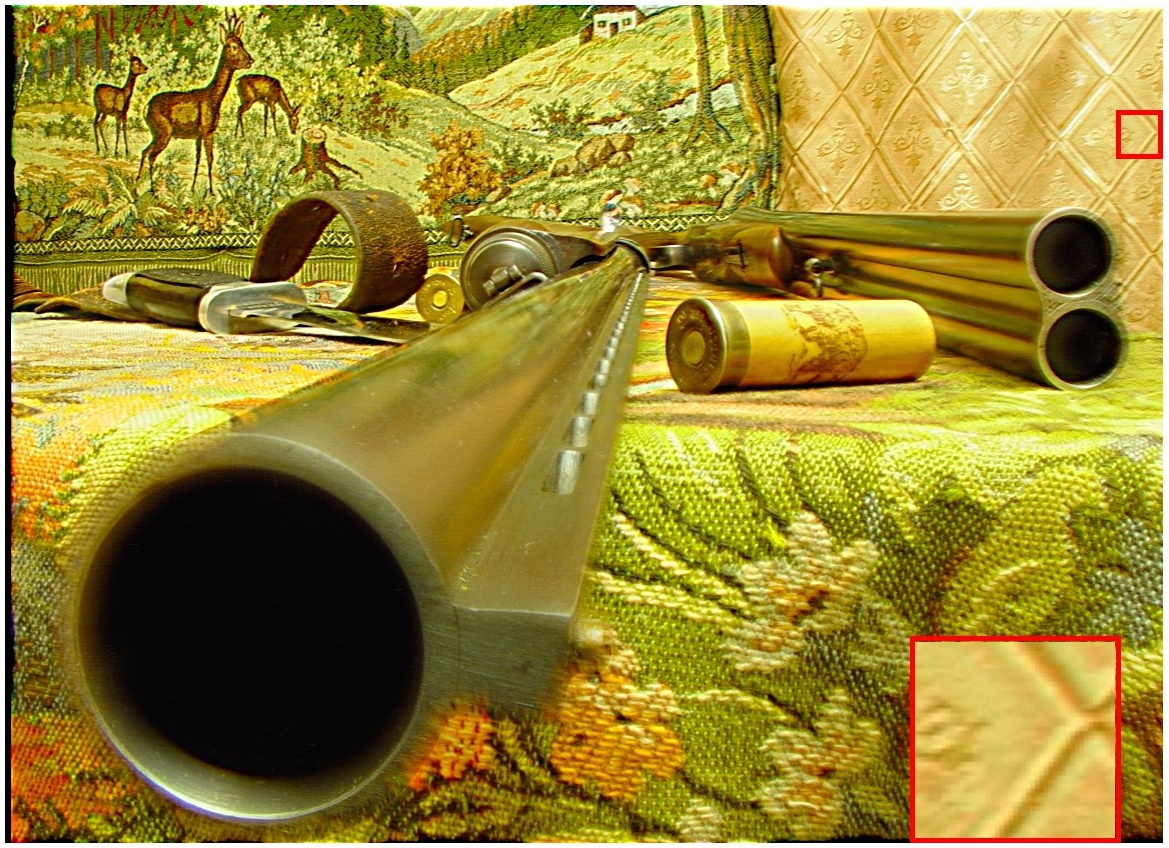}}\\
\caption{First, middle, and last frames of the perspective shifts rendered with the method proposed in \cite{Levin2010}.}
\label{fig:Levin}
\vspace{-10pt}
\end{figure*}

\subsection{Perspective shifts}
\label{ssec:exp_pers_shift}

We finally demonstrate one of the most compelling applications of the proposed light field reconstruction algorithm, which is the rendering of perspective shifts from a fixed-camera focal stack. We first use the focal stack to construct the $(x,u)$-epipolar image corresponding to each $y$ value as discussed in Section \ref{ssec:epi_im_recon}. We then determine a reference horizontal line (corresponding to a fixed $u_0$ value) on the epipolar plane and extract this  horizontal line from each $(x,u)$-epipolar image. Piling up these lines, we obtain a novel image of the scene, which is an all-in-focus image. Since each $u_0$ value on the epipolar plane represents a different viewing angle, varying the $u_0$ value of the reference horizontal lines yields a series of images that are rendered from different perspectives. 

 Fig.~\ref{persp} shows the first, intermediate and last images rendered from varying view angles for the ``Joker'' and the ``Rifle'' data sets used in Fig.~\ref{dm_ext_Nikon_rifle}. In order to test the performance of the masked back-projection algorithm without the influence of the depth map estimation stage, the results for the Joker data set presented in the top row of Fig.~\ref{persp} are obtained with a ground truth depth map of the scene. The perspective shifts in the middle and bottom rows of Fig.~\ref{persp} are rendered by using the depth maps computed with the proposed method. The results show that the proposed masked back-projection algorithm for partial light field reconstruction allows the synthesis of perspective shifts of quite competing quality. The erroneous regions observable in the reconstructed images for Joker in the middle row are due to the errors in the computation of the depth map, which are corrected when the ground truth depth map is used. The videos demonstrating the perspective shifts are available for download at the  http://www.irisa.fr/temics/demos/lightField/LightFieldPartialReconstruction.html. {Fig.~\ref{fig:Levin} shows the perspective shifts rendered for the same data sets with  the method in \cite{Levin2010}\footnote{The code made available by the authors of \cite{Levin2010} is used when testing this method. The maximum slope parameter and the PSF radius parameter in the code are manually optimized to achieve the best results, which are respectively set as 0.3 and 73 for the Joker sequence, and 1.2 and 33 for the Rifle sequence. The images presented in Fig.~\ref{fig:Levin} are rendered by setting the perspective shift parameter as $u=-50, 0, 60$ for the Joker sequence, and $u=-7, 0, 7$ for the Rifle sequence.}. Several artifacts can be observed in both sequences. The large perspective shifts in the top row of Fig.~\ref{fig:Levin} lead to significant blurring in the objects close to image borders (the melon and the king). Also, several regions close to the object boundaries in both data sets  become transparent (e.g., the left side of the melon covering the joker's hand in the top row, and the left side of the tip of the rifle barrel in the bottom row). These artifacts can be explained with the violation of the assumption in \cite{Levin2010} that the range of in-focus depth values covered by the focal stack is much larger than the depth range of the actual objects in the scene. This assumption is necessary for the reliability of the technique proposed in \cite{Levin2010}, which is based on the computation of a shifted average of all focal stack images and then deconvolving the average image to obtain the perspective shift (the loss of content along image borders in the bottom row of Fig.~\ref{fig:Levin} is due to this deconvolution). Since both focal stacks used in these experiments are such that the first image of the stack focuses on the frontmost object in the scene, the above assumption fails and the averaging of the focal stack images before deconvolution leads to the observed see-through effect. The presence of more severe artifacts in the Joker data set compared to the Rifle data set in Fig.~\ref{fig:Levin} may be due to the relatively small number of images in the focal stack (only 5 for Joker and 24 for Rifle). Our method does not suffer from these image capturing conditions and it achieves an accurate reconstruction thanks to the masked back projection technique, which exploits only the focal stack images where a region of interest is in focus in the reconstruction of that region.


\section{Conclusions and perspectives}
\label{sec:conclusion}

A static light field is a four-dimensional representation of light emitted by a scene. The recorded flow of rays contains a lot of information on the captured scene, enabling, by appropriate computation, advanced image manipulations such as digital refocusing, perspective shifts, as well as depth map estimation and saliency detection. Plenoptic cameras have recently become commercially available with however severe limitations in terms of resolution, which is far from the resolution achieved by traditional 2D cameras.
In this paper, we have described a method to reconstruct relevant parts of the light field by reconstructing epipolar images of the light field from a stack of images taken with a conventional (high resolution) camera at different focused depths. Exploiting the fact that a photograph is a two-dimensional projection of the light field along a direction that changes with the focus, we reconstruct the epipolar images of the light field by back-projecting the photographs, following principles of 4D tomography. A focus map is first estimated from all the images of the focal stack, which is then converted into a depth map with the help of a prior calibration of the camera. From the focus map, we also derive a mask locating the positions of the in-focus pixels for each image of the stack. The epipolar images are then reconstructed by back-projecting only in-focus pixels of every image of the focal stack. The angles of back-projection are estimated from the depth map. 
The proposed method allows us to render puzzling perspective shifts, extended depth of field photographs, as well  as ``click-and-refocus" features. It opens new perspectives for high resolution light field rendering. With a mere update of the software of a conventional camera that would enable focus sweeping during continuous shooting, one may be able to quickly capture a full light field with a high spatial resolution as well as a high angular resolution.\\

\appendix

\normalsize

Here we discuss the calibration of the camera to determine the focal distance. The DigicamControl software provides a focus parameter $ F_{Param}$ that is associated with the focal distance. We establish a correspondence between $ F_{Param}$ and the focus distance for a given focal length by first manually setting the zoom (focal length), then turning the focus ring to the far end until the camera is focused on the nearest objects, where $F_{Param}=0$. Next, we launch the DigicamControl software, vary $ F_{Param}$ and measure the focal distance for each value by recording the interval of distance within which a printed grid can be seen without any blur. The measured correspondence between the focus parameter and the focal distance interval is presented in Table \ref{tab:focusparam_Digicam}. We finally fit a curve between the mean values of the focused intervals and the focus parameter, which gives a one-to-one correspondence between the focus parameter and the focal distance.

%

\begin{table}[t]
\scriptsize
\centering
\begin{tabular}{|c|c|}
  \hline
  Focus Parameter & Focused Interval (in metres) \\
  \hline
 0 & 0.24-0.25 \\
  -500 & 0.27-0.28 \\
  -1000 & 0.30-0.32 \\
  -1500 & 0.35-0.37 \\
  -2000 & 0.41-0.43 \\
  -2500 & 0.49-0.51 \\
  -3000 & 0.60-0.63 \\
  -3500 & 0.79-0.82 \\
  -4000 & 1.20-1.27 \\
  -4500 & 2.0-2.7 \\
  -5000 & 15-25 \\
  \hline
\end{tabular}
\caption{Variation of focused distance with focus parameter $F_{Param}$}
\label{tab:focusparam_Digicam}
\end{table}

%

{\small
\bibliographystyle{ieee}
\bibliography{egbib}
}

\end{document}